\documentclass{article} 
\usepackage{iclr2025_conference,times}

\usepackage{amsmath,amsfonts,bm}

\def\eqref#1{equation~\ref{#1}}

\def\1{\bm{1}}

\DeclareMathAlphabet{\mathsfit}{\encodingdefault}{\sfdefault}{m}{sl}
\SetMathAlphabet{\mathsfit}{bold}{\encodingdefault}{\sfdefault}{bx}{n}

\usepackage{hyperref}
\usepackage{url}
\usepackage{amsthm}
\usepackage{hyperref}

\usepackage[most]{tcolorbox}
\usepackage{wrapfig}
\usepackage{multirow}
\usepackage{graphicx,xcolor,float}
\usepackage{subfigure}
\usepackage{threeparttable}
\usepackage[ruled,vlined]{algorithm2e}
\usepackage{colortbl}
\usepackage{color}
\usepackage{multirow}
\usepackage{tabularx}
\usepackage{float}
\usepackage{graphicx}
\usepackage{booktabs}
\usepackage{arydshln}
\usepackage{enumitem}
\usepackage{wrapfig}
\usepackage{caption}
\usepackage{graphicx}
\usepackage{makecell}
\usepackage{float} 
\usepackage{mathtools}
\usepackage{subfigure}
\usepackage{bbding}
\usepackage{makecell}
\usepackage{tabularx}
\usepackage{amssymb,mathrsfs,amsmath}
\usepackage{pifont}
\usepackage{xcolor}
\usepackage{bbm}

\usepackage{fontawesome5}

\usepackage{listings}

\newcommand{\ourmethod}{{\fontfamily{lmtt}\selectfont \textbf{LatentEvolve}}\xspace}
\newcommand{\llmname}[1]{{\fontfamily{pcr}\selectfont {#1}}\xspace}

\definecolor{bittersweet}{rgb}{1.0, 0.44, 0.37}
\definecolor{mygreen}{rgb}{0.29, 0.7, 0.48}
\usepackage{pifont}
\definecolor{demphcolor}{RGB}{144,144,144}

\definecolor{mygray}{gray}{0.4}
\definecolor{autopurple}{HTML}{7030A0}
\definecolor{dyna_yellow}{HTML}{BF9000}
\definecolor{adaptive_blue}{HTML}{0070C0}
\definecolor{darksalmon}{rgb}{0.91, 0.59, 0.48}
\definecolor{emerald}{rgb}{0.31, 0.78, 0.47}
\definecolor{green(pigment)}{rgb}{0.0, 0.65, 0.31}
\definecolor{amaranth}{rgb}{0.9, 0.17, 0.31}
\definecolor{iris}{rgb}{0.35, 0.31, 0.81}
\definecolor{uu}{rgb}{0.95, 0.51, 0.51}
\definecolor{spirodiscoball}{rgb}{0.06, 0.75, 0.99}
\definecolor{mygrey}{gray}{0.4}

\usepackage{svg}
\usepackage{twemojis}

\usepackage[table,xcdraw,usenames,dvipsnames]{xcolor}
\usepackage{cleveref}
\SetKwInOut{Input}{Input}\SetKwInOut{Output}{Output}

\definecolor{QuestionColor}{rgb}{0.7, 0.1, 0.1} 
\definecolor{AnswerColor}{rgb}{0.1, 0.5, 0.1} 
\definecolor{ReasoningColor}{rgb}{0.1, 0.1, 0.7} 

\hypersetup{
    colorlinks=true,
    linkcolor=red,
    citecolor=cyan,
    filecolor=magenta,      
    urlcolor=magenta,
    }

\title{LatentEvolve: Self-Evolving Test-Time Scaling in Latent Space}

\author{Guibin Zhang$^{\twemoji{lion}\dag}$, 
Fanci Meng$^{\twemoji{bird}\dag}$, 
Guancheng Wan$^{\twemoji{bear}}$, 
\textbf{Zherui Li}$^{\twemoji{lion}}$,\\
\textbf{Kun Wang}$^{\twemoji{deer}}$, 
\textbf{Zhenfei Yin}$^{\twemoji{goggles}}$, 
\textbf{Lei Bai}$^{\twemoji{goggles}}$, 
\textbf{Shuicheng Yan}$^{\twemoji{lion}}$ \\
	$^{\twemoji{lion}}$NUS\quad
    $^{\twemoji{bird}}$USTC \quad
    $^{\twemoji{bear}}$UCLA \quad
    $^{\twemoji{deer}}$NTU \quad
    $^{\twemoji{goggles}}$Shanghai AI Lab\quad $^\dag$Equal Contribution\\
{\faEnvelope} {Main Contact}: \texttt{guibinz@outlook.com}
}

\iclrfinalcopy 
\begin{document}

\maketitle

\begin{abstract}
Test-time Scaling (TTS) has been demonstrated to significantly enhance the reasoning capabilities of Large Language Models (LLMs) during the inference phase without altering model parameters. However, existing TTS methods are largely independent, implying that LLMs have not yet evolved to progressively learn how to scale more effectively. With the objective of evolving LLMs to learn ``how to scale test-time computation,'' we propose \ourmethod, a self-evolving latent TTS framework inspired by the complementary learning system (CLS) theory. Analogous to the human brain's dual system of a fast-recall hippocampus and a slow-consolidating neocortex, \ourmethod comprises two evolutionary components: \textit{daytime scaling}, which rapidly retrieves historical latent representations to better guide current LLM reasoning; and \textit{nighttime scaling}, which integrates past latent optimizations in a manner akin to the human brain's consolidation of experiences during sleep. The alternation of daytime and nighttime processes facilitates a fast and slow evolution of LLM TTS, mirroring human cognitive dynamics in a fully unsupervised manner. Extensive experiments across eight benchmarks and five model backbones demonstrate that our \ourmethod surpasses state-of-the-art TTS methods such as LatentSeek and TTRL by up to $13.33\%$ and exhibits exceptional cross-domain and cross-backbone generalization. The codes are available at \url{https://github.com/jins7/LatentEvolve}.
\end{abstract}

\vspace{-0.4em}
\section{Introduction}

\vspace{-0.6em}
The general capabilities of large language models (LLMs) have been extensively developed and widely recognized across numerous domains, such as mathematical reasoning~\citep{zeng2024skyworkmathdatascalinglaws,wu2025agenticreasoningreasoningllms}, software engineering~\citep{wei2025swerladvancingllmreasoning,deepswe2025,yang2024sweagentagentcomputerinterfacesenable}, multimodal understanding~\citep{zheng2025deepeyesincentivizingthinkingimages,su2025pixelreasonerincentivizingpixelspace}, and embodied action~\citep{wang2023voyageropenendedembodiedagent}, emerging as dominant paradigms that are steadily advancing toward artificial general intelligence (AGI)~\citep{bubeck2023sparks}. Much of this success in recent years has been driven by \textit{training-time scaling}, wherein increasing the volume of training data and parameters consistently yields performance improvements~\citep{kaplan2020scalinglawsneurallanguage,aghajanyan2023scalinglawsgenerativemixedmodal}. However, the pace of this scaling, particularly in terms of pre-training scale, has begun to slow, constrained by its resource-intensive nature and the depletion of high-quality training data~\citep{villalobos2022will,zhou2025surveyllmtimesdata}. Consequently, a growing body of research has shifted attention to \textit{test-time scaling} (TTS)~\citep{zhang2025surveytesttimescalinglarge,chung2025revisitingtesttimescalingsurvey}, aiming to fully harness the intrinsic knowledge of LLMs to maximize their real-world utility without additional training during the test phase.

The dimensions of TTS are highly diverse. One prominent form is \textbf{(I) parallel scaling}, wherein multiple candidate responses are generated for a given query, which are subsequently aggregated via an appropriate mechanism. This can involve multiple samples from a single LLM~\citep{brown2024largelanguagemonkeysscaling,snell2024scalingllmtesttimecompute} or sampling from multiple heterogeneous LLMs~\citep{zhang2025avengers,ye2025multiagentsamplingscalinginference}. Another form is \textbf{(II) sequential scaling}, where the LLM iteratively refines solutions based on its own previous outputs, and which underlies many ``System~2''-style generation methods~\citep{yu2024distilling21,wei2023cot,he2024retrievingrethinkingrevisingchainofverification,gou2024criticlargelanguagemodels}. Other variants include \textit{hybrid} approaches that integrate both strategies~\citep{wang2024mixture,graphofthoughts}, as well as \textit{internalized scaling}, where models like DeepSeek~R1~\citep{guo2025deepseek} and OpenAI~o-series~\citep{li2025searcho1agenticsearchenhancedlarge} are inherently capable of adaptively allocating computational resources during inference.

However, regardless of the specific form, most TTS paradigms lack the capacity for \textit{self-evolution}, as inference-time computations for distinct queries are typically treated as mutually \textbf{independent} events. For example, in verbal reinforcement learning approaches such as Reflexion~\citep{shinn2023reflexionlanguageagentsverbal} and Mind Evolution~\citep{lee2025evolvingdeeperllmthinking}, successful reflective strategies are instance-specific and are not transferred to subsequent tasks. Likewise, in ``sampling-and-voting'' scaling methods~\citep{brown2024largelanguagemonkeysscaling,irvine2023rewardingchatbotsrealworldengagement}, prior successes in selecting the correct answer do not inform or refine future selection strategies. This \textit{inter-task independence} fundamentally constrains the potential of TTS paradigms to progressively evolve through continual interaction with the environment. This raises a natural yet critical research question: \textit{How can we design a TTS framework that learns from experience, enabling its scaling capabilities to evolve and improve as it solves more problems?}

To address this challenge, we introduce \ourmethod, a self-evolving TTS framework inspired by the Complementary Learning Systems (CLS) theory~\citep{mcclelland1995there,kumaran2016learning}. CLS theory posits that the brain uses two synergistic systems: a fast-learning hippocampus for specific episodic memories, and a slow-learning neocortex for consolidating these experiences into general knowledge. Analogously, \ourmethod operates through a dual-phase evolution:

\vspace{-0.5em}
\begin{itemize}[leftmargin=2em,itemsep=-0.1em]
\item[\faSun] \textbf{{Daytime} Scaling} for \textit{fast, episodic adaptation}: For each new query, \ourmethod performs instance-level latent optimization that steers the LLM toward better reasoning paths. This process is initialized by retrieving relevant ``episodic traces'', \textit{i.e.}, latent representations from previously solved problems, mirroring the \textit{daytime} fast recall of individual memories.

\item[\faCloudMoon] \textbf{{Nighttime} Scaling} for \textit{slow, procedural consolidation}: Mirroring how the brain consolidates experiences into general skills during sleep, \ourmethod periodically fine-tunes a compact knowledge consolidation model (\textit{latent weaver}) on the collection of \textit{daytime} traces. This \textit{nighttime} process distills these specific experiences into procedural knowledge, evolving to generate superior initial latent representations for future tasks.

\end{itemize}

\begin{figure}
\centering
\includegraphics[width=0.9\textwidth]{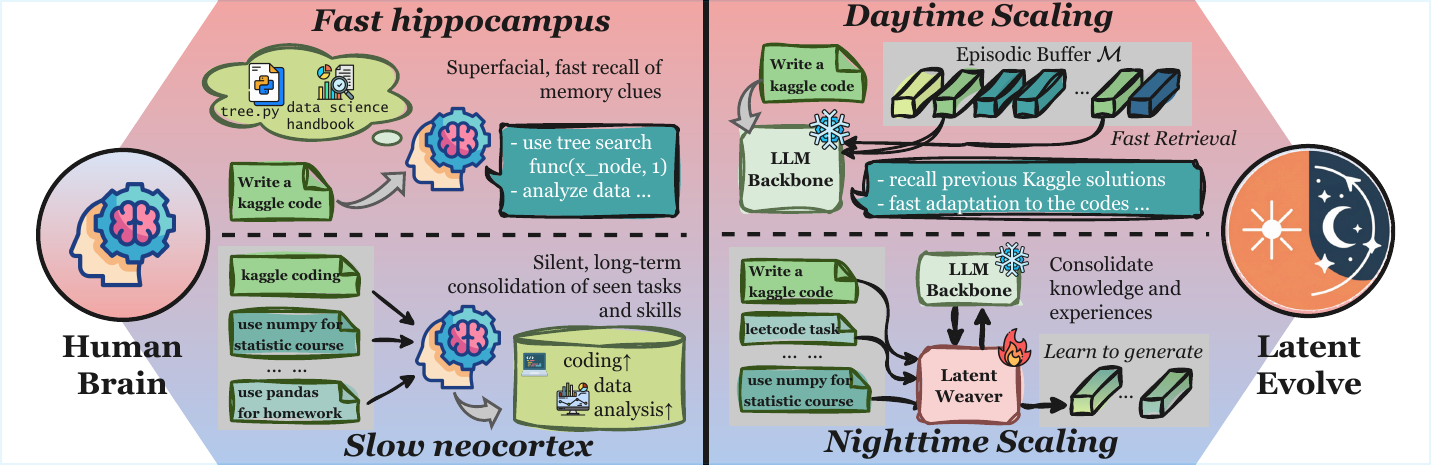}
\vspace{-0.8em}
\caption{The daytime scaling of \ourmethod functions analogously to the human hippocampus, rapidly retrieving memory cues, whereas the nighttime scaling mirrors the neocortex during sleep, performing deep integration of accumulated knowledge.} \label{fig:intro}
\vspace{-1.5em}
\end{figure}

Within this continual interplay, \ourmethod enables LLMs to perform test-time computation during daytime inference while simultaneously accumulating experiential knowledge. During nighttime reflection, these experiences are periodically consolidated into endogenous procedural memory, thereby achieving a ``fast-slow'' evolution of test-time scaling. The entire process operates \textbf{without} reliance on ground-truth labels or any other external signals.

\vspace{-0.4em}
\paragraph{Experimental Observation.}  
Extensive evaluations across eight benchmarks spanning four domains demonstrate that \ourmethod provides:  
\ding{182} \textbf{high performance}: achieving up to $23.3\%$ gains on math reasoning, surpassing GRPO and LatentSeek on MATH-500 by $1.75\%$ and $11.40\%$, respectively;  
\ding{183} \textbf{cross-domain generalization}: test-time scaling on MMLU and MATH transfers to out-of-domain datasets, yielding gains of $7.07\%$ on GPQA and $5.22\%$ on JAMA;  
\ding{184} \textbf{continual learning ability}: test-time scaling across multiple new domains does not degrade performance on previously seen domains and can even provide modest improvements.

\vspace{-0.4em}
\section{Related Work}
\vspace{-0.6em}
\textbf{Test-time computation} is a canonical pathway for transitioning from System~1 to System~2 models, with two primary branches: \textit{test-time training} (TTT) and \textit{test-time scaling} (TTS). The former involves updating model parameters during the test phase in an unsupervised manner, as exemplified by TTT~\citep{sun2020test,akyurek2024surprisingttt}, TTT+++~\citep{liu2021ttt++}, and SIFT~\citep{hubotter2025efficientlysift}. The latter increases computational expenditure without altering parameters, which can occur in the \textbf{(I) explicit, natural-language space}, as in self-correction~\citep{shinn2023reflexionlanguageagentsverbal,gou2024criticlargelanguagemodels,kang2025t1toolintegratedselfverificationtesttime}, feedback modeling~\citep{cobbe2021trainingverifier,yu2025selfgeneratedcritiquesboostreward}, or repeated sampling~\citep{gui2024bonbonalignmentlargelanguage,ye2025multiagentsamplingscalinginference}; or it may operate in the \textbf{(II) latent space}, where methods such as Coconut~\citep{hao2024coconuttraininglargelanguagemodels} and SoftCoT~\citep{xu2025softcotsoftchainofthoughtefficient,xu2025softcottesttimescalingsoft} perform deep scaling within the model’s hidden representations. Our proposed \ourmethod falls primarily within the latent TTS. Yet, regardless of form, existing approaches are rarely capable of rapid evolution through the ongoing process of problem solving, a limitation that \ourmethod is designed to overcome.

\vspace{-0.6em}
\paragraph{Latent Computation \& Reasoning} seeks to exploit continuous latent representations, rather than discrete language space, to enable a more machine-native and concise form of reasoning for LLMs~\citep{zhu2025surveylatentreasoning}. Mainstream approaches can be broadly categorized as: \textbf{(I) architecturally enabling native latent reasoning}, as exemplified by Coconut~\citep{hao2024coconuttraininglargelanguagemodels}, CoLaR~\citep{tan2025thinksilentlythinkfast}, and Recurrent Depth~\citep{geiping2025scalingtesttimecomputelatent}; and \textbf{(II) employing latent computation to steer LLM generation}, as in LatentSeek~\citep{li2025seekdarkreasoningtesttime}, SoftCoT~\citep{xu2025softcottesttimescalingsoft,xu2025softcotsoftchainofthoughtefficient}, and otherss~\citep{liu2024deliberation,sun2025enhancinglatentcomputationtransformers}, which leverage latent representations as an intervention to modulate the quality of generated outputs. Other methods, such as IMM~\citep{orlicki2025wordslatentmemoryapproach} and MemoryLLM~\citep{wang2024memoryllm,wang2025mextendingmemoryllmscalable}, employ latent tokens as a means of preserving contextual memory. Distinct from these approaches, \ourmethod implements a dual-stage test-time evolution within the latent space, whereas prior strategies generally remain inter-task independent.

\textbf{Self-Evolving LLM \& Agent.} How to evolve LLMs during their interactions with the environment has drawn increasing attention from the research community~\citep{gao2025surveyselfevolvingagentspath,fang2025comprehensivesurveyselfevolvingai}. Existing approaches generally employ certain carriers for evolution, including: \textbf{(I) parametric update}, wherein prior experiences are encoded directly into model parameters~\citep{zeng2023agenttuningenablinggeneralizedagent,chen2024agentflandesigningdatamethods,zhao2025absolutezeroreinforcedselfplay,chen2025selfevolvingcurriculumllmreasoning}; \textbf{(II) experience databases}, in which past problem-solving trajectories~\citep{zhao2024expelllmagentsexperiential,song2024agentbankgeneralizedllmagents} or distilled experiential knowledge~\citep{zhang2025gmemorytracinghierarchicalmemory,wang2025mobileagenteselfevolvingmobileassistant,tang2025agentkbleveragingcrossdomain} are leveraged to contextually enhance LLM capabilities; and \textbf{(III) skill condensation}, where reusable tools (\textit{e.g.}, APIs, MCPs) are encapsulated as functional assets~\citep{zheng2025skillweaverwebagentsselfimprove,suzgun2025dynamiccheatsheettesttimelearning,zhang2025darwingodelmachineopenended,qiu2025alitageneralistagentenabling,qiu2025agentdistilltrainingfreeagentdistillation}. Distinct from these paradigms, \ourmethod performs test-time evolution within the latent space, treating the latent sequences as a compact and adaptable skill repository.

\section{Preliminary}
\vspace{-0.4em}
In this section, we formally describe the procedure of current latent-based TTS methods, which manage to steer LLM's generative process by introducing adaptable, continuous vectors.

\vspace{-0.4em}
\paragraph{Latent-Space Aided Reasoning.}
Let $\pi_{\boldsymbol{\theta}}$ be a language model with frozen parameters $\boldsymbol{\theta}$. For a given problem context $\mathbf{c}$, the standard generative process produces an output sequence $\mathbf{y}$ by sampling from the conditional probability distribution $p(\mathbf{y}|\mathbf{c}; \boldsymbol{\theta})$.
The core principle of this paradigm is to introduce an auxiliary sequence of continuous vectors, $\mathbf{z} = ({z}_1, {z}_2, \cdots, {z}_L)$, which we refer to as a \textit{latent token sequence}. These vectors act as a dynamic, instance-specific control signal that conditions the generative process of the frozen LLM. The generation is thus reformulated as sampling from a new distribution, conditioned on both the original context and the latent intervention:
\begin{equation}
\mathbf{y} \sim p(\mathbf{y} | \mathbf{c}, \mathbf{z}; \boldsymbol{\theta})
\label{eq:latent_intervention}
\end{equation}
The latent sequence $\mathbf{z}$ can be introduced through various mechanisms, such as being prepended to input embeddings, directly augmenting the model's internal key-value (KV) cache, or representing a latent thought process for subsequent decoding~\citep{xu2025softcottesttimescalingsoft,liu2024deliberation,sun2025enhancinglatentcomputationtransformers}.
The primary objective is to find an optimal latent intervention $\mathbf{z}^*$ that maximizes an objective function $J(\mathbf{z})$. This objective is formalized as the expected quality of the generated output:
\begin{equation}
\mathbf{z}^* = \arg\max_{\mathbf{z}} J(\mathbf{z}), \;\; \text{where} \;\; J(\mathbf{z}) = \mathbb{E}_{\mathbf{y} \sim p(\mathbf{y}|\mathbf{c}, \mathbf{z}; \boldsymbol{\theta})} [Q(\mathbf{y})]
\label{eq:objective}
\end{equation}
where $Q(\mathbf{y})$ is a scoring function that evaluates the quality of an output $\mathbf{y}$.

\vspace{-0.4em}
\paragraph{Generation of Latent Representations.}
The mechanism for generating the latent sequence $\mathbf{z}$ defines the specific TTS method. Existing work either optimizes a single set of \textit{task-specific} soft prompts, $\mathbf{z}_{\text{task}}$, applied across all instances~\citep{xiao2023decomposedprompttuninglowrank,choi-etal-2023-smop}, or performs \textit{query-specific} optimization to find a bespoke latent path $\mathbf{z}_i$ for each individual query $\mathbf{c}_i$~\citep{li2025seekdarkreasoningtesttime,peng2024softprompttuningaugmenting,xu2025softcottesttimescalingsoft,sun2025enhancinglatentcomputationtransformers}. Whatever the granularity is, most of these practices are self-contained, \textit{i.e.}, do not rapidly learn or evolve from one instance to the next, thereby incapable of on-the-fly adaptation based on cumulative experience.

\begin{figure}
\includegraphics[width=\textwidth]{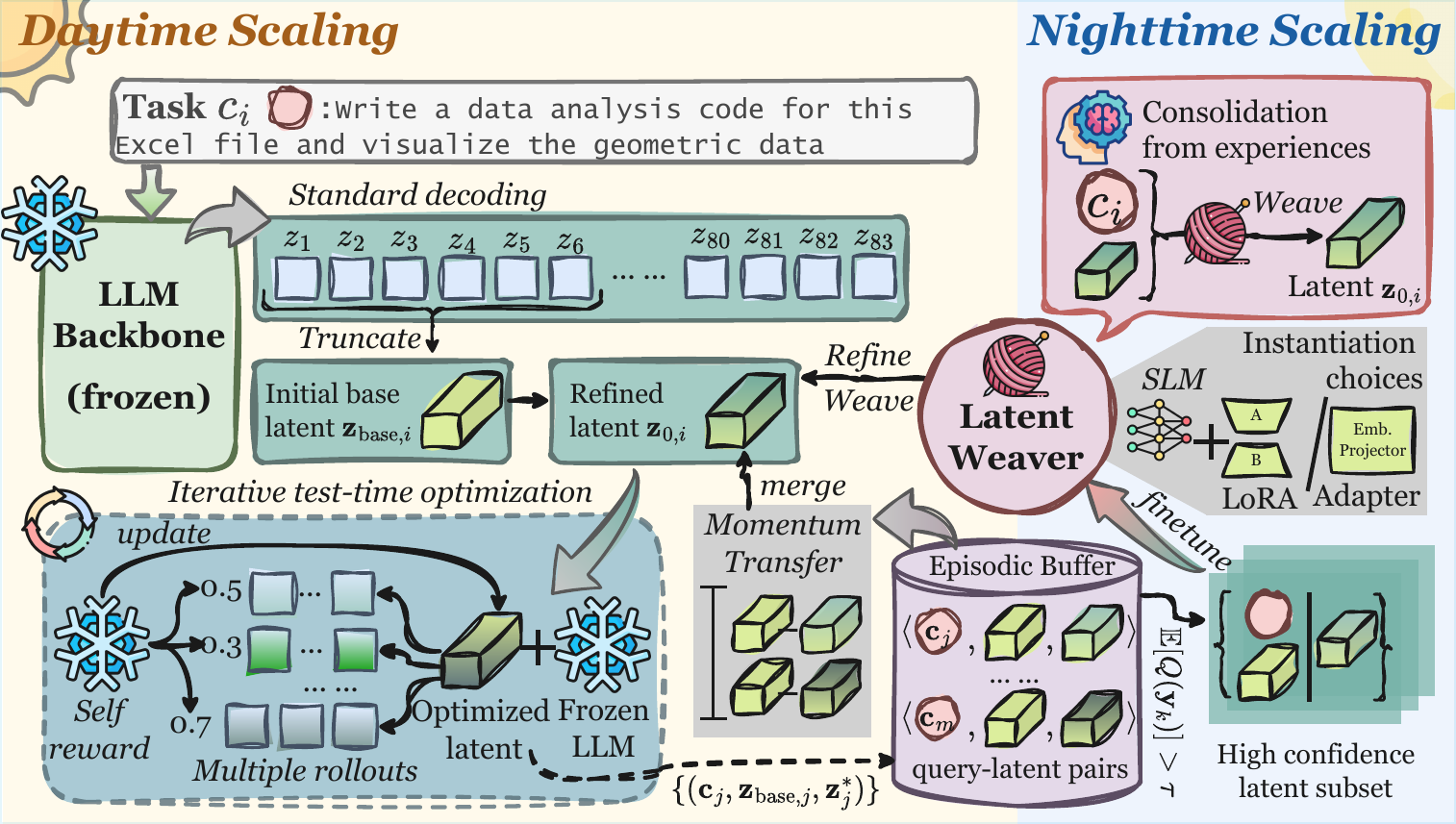}
\vspace{-1.5em}
\caption{The overview of our proposed \ourmethod.} \label{fig:framework}
\vspace{-1.5em}
\end{figure}

\section{Methodology}

\vspace{-0.5em}
\ourmethod unfolds as a dual-phase evolving process that enables LLMs to adapt and self-improve at test time. First, we introduce \textit{daytime test-time scaling} ($\triangleright$ \Cref{sec:daytime}), which performs fast, instance-specific adaptation guided by weighted momentum transfer. Then, \textit{nighttime deliberative consolidation} ($\triangleright$ \Cref{sec:nighttime}) integrates these episodic traces into a compact parametric prior through the latent weaver. Finally, \textit{dual-phase evolving scaling} ($\triangleright$ \Cref{sec:dual-phase}) ties the two phases into a recurrent cycle, ensuring continual interleaved evolution in latent space.

\vspace{-0.5em}
\subsection{Daytime Test-time Optimization}\label{sec:daytime}
\vspace{-0.5em}
The \textit{Daytime Scaling} is designed for fast, on-the-fly adaptation, mirroring the brain's ability to rapidly recall specific past experiences to navigate a present challenge. This process unfolds in three key stages for each incoming query: retrieving relevant memories, constructing an informed initial latent sequence, and refining it through self-guided optimization.

\vspace{-0.5em}
\paragraph{Associative Retrieval.}
Inspired by the function of episodic memory in cognitive science, \ourmethod maintains an \textbf{episodic buffer}, $\mathcal{M}$, which serves as a dynamic archive of specific, high-quality test-time scaling experiences. Each entry is a triplet $(\mathbf{e}_{\mathbf{c}_j}, \mathbf{z}_{\text{base},j}, \mathbf{z}^*_j)$, storing a previous query's context embedding $\mathbf{e}_{\mathbf{c}_j}$, its initial latent sequence $\mathbf{z}_{\text{base},j}$, and its refined latent sequence $\mathbf{z}^*_j$.

Upon receiving a new query, which we define as the input prompt $\mathbf{c}_i$, we first compute its semantic embedding $\mathbf{e}_{\mathbf{c}_i}$ using the frozen LLM's final hidden state. We then perform a similarity search to retrieve a neighborhood of the top-$k$ most relevant experiences from the buffer:
\begin{equation}
\mathcal{N}_k(\mathbf{c}_i) = \text{Top-k}_{j} \{ (\mathbf{e}_{\mathbf{c}_j}, \mathbf{z}_{\text{base},j}, \mathbf{z}^*_j) \in \mathcal{M} \}, \quad \text{based on similarity } S(\mathbf{e}_{\mathbf{c}_i}, \mathbf{e}_{\mathbf{c}_j}),
\end{equation}
where $S(\cdot, \cdot)$ is instantiated via cosine similarity. This allows the upcoming test-time optimization to benefit from a small, highly relevant subset of its past experiences.

\vspace{-0.4em}
\paragraph{Informed Latent Initialization.}  
A well-informed starting point can substantially improve both the efficiency and quality of reasoning. For each query $\mathbf{c}_i$, we first derive a base initialization $\mathbf{z}_{\text{base},i}$ via an initial Chain-of-Thought (CoT) decoding, taking the prefix of the resulting latent sequence:
\begin{equation}
\mathbf{z}_{\text{base},i} = H_{\boldsymbol{\theta}}(\mathbf{c}_i)_{1:L'}
\label{eq:base_init}
\end{equation}
where $H_{\boldsymbol{\theta}}(\mathbf{c}_i)$ denotes the full latent sequence produced by $\pi_{\boldsymbol{\theta}}$ under greedy decoding, and the subscript $1:L'$ selects the first $L'$ latent vectors. This base state serves as a preliminary reasoning trajectory, which can be further refined using the retrieved neighborhood $\mathcal{N}_k(\mathbf{c}_i)$ to form a superior initialization $\mathbf{z}_{0,i}$.  A naive approach might be to simply average the retrieved final latent sequences $\mathbf{z}^*_j$, but this can be misleading as different queries may yield conflicting patterns. Instead, we follow a more intuitive principle: it is not the final latent states that matter most, but the journey from the initial $\mathbf{z}_{\text{base},j}$ to the refined $\mathbf{z}^*_j$. We capture this journey as the optimization ``momentum'', $\Delta\mathbf{z}_j = \mathbf{z}^*_j - \mathbf{z}_{\text{base},j}$, and introduce \textbf{weighted momentum transfer}: 
By aggregating these momenta weightedly, we guide $\mathbf{z}_{\text{base},i}$ toward regions of the latent space that have been fruitful in the past:
\begin{equation}
\mathbf{z}_{0,i} = \mathbf{z}_{\text{base},i} + \sum_{j \in \mathcal{N}_k(\mathbf{c}_i)} \alpha_j \Delta\mathbf{z}_j, \;\;\text{where}\;\; \alpha_j \propto \exp(S(\mathbf{e}_{\mathbf{c}_i}, \mathbf{e}_{\mathbf{c}_j})).
\label{eq:momentum}
\end{equation}
In this way, the initialization is gently steered not only toward promising regions but also along trajectories that have historically led to better outputs, allowing reasoning to begin with a well-informed and contextually grounded foundation.

\vspace{-0.5em}
\paragraph{Self-Supervised Refinement and Archiving.}
Although the informed initial state $\mathbf{z}_{0,i}$ offers a promising foundation, it is not tailored to the specific context $\mathbf{c}_i$ and thus requires refinement to enhance reasoning performance. We adopt a self-rewarding strategy, a paradigm broadly validated in prior work~\citep{li2025seekdarkreasoningtesttime,yuan2025selfrewardinglanguagemodels,zuo2025ttrltesttimereinforcementlearning}. Concretely, the LLM $\pi_\theta$ serves as its own evaluator by assigning a quality score $Q(\mathbf{y}_k)$ to the output $\mathbf{y}_k$ generated under the guidance of $\mathbf{z}_{0,i}$ (see the detailed implementation of $Q(\cdot)$ in \Cref{app:refine-latent}). The latent sequence is then iteratively refined through gradient ascent with respect to this self-supervised signal. The gradient of $J(\mathbf{z}_k)$ is estimated via policy gradient~\citep{williams1992simple} as:
\begin{equation}\label{eq:daytime-optimize-loss}
\small
\nabla_{\mathbf{z}_k} J(\mathbf{z}_k) 
= \nabla_{\mathbf{z}_k} \mathbb{E}_{\mathbf{y}\sim p(\mathbf{y}\mid \mathbf{c}_i,\mathbf{z}_k;\boldsymbol{\theta})}[Q(\mathbf{y})]
\approx \frac{1}{M}\sum_{m=1}^M Q(\mathbf{y}^{(m)}) \nabla_{\mathbf{z}_k}\log p(\mathbf{y}^{(m)}\mid \mathbf{c}_i,\mathbf{z}_k;\boldsymbol{\theta}),
\end{equation}
where $\{\mathbf{y}^{(m)}\}_{m=1}^M$ are samples drawn from $p(\cdot \mid \mathbf{c}_i,\mathbf{z}_k;\boldsymbol{\theta})$ by $M$ times. Accordingly, the latent state is iteratively updated as $\mathbf{z}_{k+1} \leftarrow \mathbf{z}_k + \eta \nabla_{\mathbf{z}_k} J(\mathbf{z}_k)$, where $\eta$ is the learning rate. The refinement terminates either after $K$ iterations or once $\mathbb{E}[Q(\mathbf{y}_k)]$ has failed to increase for three successive rounds, yielding the final latent state $\mathbf{z}^*_i$, under whose guidance $\pi_\theta$ produces the ultimate output $\mathbf{y}$. The triplet $(\mathbf{e}_{\mathbf{c}_i}, \mathbf{z}_{\text{base},i}, \mathbf{z}^*_i)$ is archived into $\mathcal{M}$ whenever $\mathbb{E}[Q(\mathbf{y}_k)]$ exceeds a predefined threshold $\tau$ (see detailed process in \Cref{app:night}). Thus, the preservation of high-confidence experiences deepens the repository from which \ourmethod continually distills its evolving knowledge.

\vspace{-0.3em}
\subsection{Nighttime Deliberative Consolidation}\label{sec:nighttime}
\vspace{-0.5em}
While the \textit{daytime scaling} excels at rapid, instance-level adaptation, its knowledge remains fragmented within the buffer. To achieve generalizable improvement, these scattered experiences must be integrated into a coherent procedural skill, which is also the purpose of the \textit{nighttime scaling}, analogous to the neocortex's role in consolidating memories into abstract knowledge during sleep.

\vspace{-0.5em}
\paragraph{Latent Weaver.}
To perform this consolidation, we introduce the \textit{latent weaver} $\mathbf{W}_{\boldsymbol{\psi}}$, aimed at distilling the collective wisdom from the episodic buffer. Technically, $\mathbf{W}_{\boldsymbol{\psi}}$ is trained to predict the refined latent sequence $\mathbf{z}^*_j$ conditioned on the context embedding $\mathbf{e}_{\mathbf{c}_j}$ and the base state $\mathbf{z}_{\text{base},j}$, thereby enabling rapid and precise test-time scaling. We instantiate $\mathbf{W}_{\boldsymbol{\psi}}$ via a smaller LLM $\psi$.

\vspace{-0.5em}
\paragraph{Consolidation through Experience Replay.}
Periodically, after the episodic buffer $\mathcal{M}$ has accumulated a sufficient number of high-confidence experiences, the nighttime consolidation is triggered. The experience triplets $\{(\mathbf{e}_{\mathbf{c}_j}, \mathbf{z}_{\text{base},j}, \mathbf{z}^*_j)\}$ from $\mathcal{M}$ are leveraged to update the parameters $\boldsymbol{\psi}$ of the latent weaver. The training objective is to minimize the reconstruction error between the weaver's prediction and the archived optimal latent sequence:
\begin{equation}
\mathcal{L}(\boldsymbol{\psi}) = \mathbb{E}_{(\mathbf{e}_{\mathbf{c}_j}, \mathbf{z}_{\text{base},j}, \mathbf{z}^*_j) \sim \mathcal{M}} \left[ \left\| \mathbf{W}_{\boldsymbol{\psi}}(\mathbf{e}_{\mathbf{c}_j}, \mathbf{z}_{\text{base},j}) - \mathbf{z}^*_j \right\|^2_2 \right],
\label{eq:consolidation_loss}
\end{equation}
which effectively \textit{weaves} the sparse, episodic optimization experiences into the continuous parametric space of the model. Through such nighttime scaling, $\mathbf{W}_\psi$ is imbued with procedural intuition and capable of generating superior initial reasoning paths for subsequent LLM reasoning. In the next section, we illustrate the overall picture of the dual-phase evolving process.

\vspace{-0.5em}
\subsection{Dual-Phase Evolving Scaling}\label{sec:dual-phase}
\vspace{-0.5em}
In this section, we formally describe the dual-phase evolution process of \ourmethod. The daytime and nighttime mechanisms, though effective in isolation, realize their full potential when embedded in a recurring cycle that mirrors the brain’s complementary learning systems: the hippocampus for rapid encoding of episodic traces and the neocortex for gradual schema formation. For each incoming query $\mathbf{c}_i$, the latent weaver $\mathbf{W}_{\boldsymbol{\psi}}$ first transforms the base latent state (except in the initial round, when $\mathbf{W}_{\boldsymbol{\psi}}$ remains untrained) to yield a refined $\mathbf{z}'_{\text{base},i}$. This is followed by daytime scaling, which, via momentum transfer, produces $\mathbf{z}_{0,i}$. Iterative refinement then generates the final latent sequence $\mathbf{z}^*_i$ through self-guided optimization, ensuring that each query benefits not only from episodic recall but also from the procedural insights accumulated during prior nighttime consolidations:
\begin{equation}
\mathbf{z}'_{\text{base},i} = \mathbf{W}_{\boldsymbol{\psi}}(\mathbf{e}_{\mathbf{c}_i}, \mathbf{z}_{\text{base},i}), \quad
\mathbf{z}^*_i = \Phi_{\text{day}}(\mathbf{c}_i, \mathbf{z}'_{\text{base},i}, \mathcal{M}; \boldsymbol{\theta}),
\label{eq:day}
\end{equation}
where $\Phi_\text{day}$ denotes daytime optimization of a given query $\mathbf{c}_i$ under the assistance of $\mathcal{M}$ and $\pi_\theta$, as described in \Cref{eq:daytime-optimize-loss}. Over time, the episodic buffer $\mathcal{M}$ accumulates triplets of adaptations $\{(\mathbf{e}_{\mathbf{c}_j}, \mathbf{z}_{\text{base},j}, \mathbf{z}^*_j)\}$. At periodic intervals (specifically, we set $T=200$ test-time instances as one cycle), nighttime scaling is invoked to consolidate accumulated experiences by updating $\mathbf{W}_{\boldsymbol{\psi}}$:
\begin{equation}
\mathbf{W}_{\boldsymbol{\psi}} \leftarrow \Phi_{\text{night}}(\mathcal{M}, \mathbf{W}_{\boldsymbol{\psi}}),
\label{eq:night}
\end{equation}
where $\Phi_{\text{night}}$ denotes experience replay and parametric distillation, as described in \Cref{eq:consolidation_loss}. The overall evolution is thus expressed as the alternating transformation
\begin{equation}
(\mathcal{M}, \mathbf{W}_{\boldsymbol{\psi}}) \;\xrightarrow{\;\;\Phi_{\text{day}}\;\;}\; \mathcal{M}' 
\;\xrightarrow{\;\;\Phi_{\text{night}}\;\;}\; (\mathcal{M}', \mathbf{W}'_{\boldsymbol{\psi}}),
\label{eq:cycle}
\end{equation}
which continually refreshes the episodic buffer while also imbuing the weaver with generalized procedural knowledge. 
This perpetual cycle of experience and consolidation allows \ourmethod's reasoning capabilities to self-evolve on the fly, entirely in an \textbf{unsupervised manner without reliance on any external labels}.

\vspace{-0.4em}
\section{Experiments}
\vspace{-0.6em}
\subsection{Experiment Setting}\label{sec:setup}
\vspace{-0.6em}
\paragraph{Backbones.} To evaluate the generalizability of \ourmethod, we experiment with LLMs from different families and of varying sizes, including \llmname{Llama-3.2-3b}~\citep{grattafiori2024llama}, \llmname{Qwen2.5-7b-instruct}~\citep{qwen2025qwen25technicalreport}, \llmname{Qwen3-4b-instruct-2507}, \llmname{Qwen3-8b}~\citep{yang2025qwen3}, and \llmname{Gemma-3-12b-it}~\citep{team2025gemma}.

\vspace{-0.7em}
\paragraph{Benchmarks.}  We conduct a comprehensive evaluation of \ourmethod across eight benchmarks from four task domains: \ding{110} \textit{general QA}, MMLU~\citep{hendrycks2021measuringmassivemultitasklanguage}; \ding{110} \textit{mathematical reasoning}, including GSM8K~\citep{cobbe2021trainingverifier}, MATH-500~\citep{hendrycks2021measuringmathematicalproblemsolving}, and AIME 2024/2025~\citep{li2024numinamath}; \ding{110} \textit{scientific reasoning}, SciBench~\citep{wang2024scibenchevaluatingcollegelevelscientific} and GPQA-Diamond~\citep{rein2023gpqagraduatelevelgoogleproofqa}; \ding{110} \textit{medical reasoning}, JAMA Clinical Challenge~\citep{chen2025benchmarkingjama}. Detailed dataset statistics are listed in \Cref{app:data}.

\vspace{-0.7em}
\paragraph{Evaluation Setup.} We apply \ourmethod independently to each benchmark's test set, except for AIME24/25 where the test size is limited, on which we evaluate after applying \ourmethod on MATH-500. We set the maximum generation length to $2048$ tokens. The small LLM used for latent weaver $\mathbf{W}_{\boldsymbol{\psi}}$ is consistently set as \llmname{Qwen-2.5-1.5b}. The dimension $L'$ in \Cref{eq:base_init} is set as $15$, the threshold $\tau$ equals $0.5$, and the dual-evolution period $T$ is $200$. The learning rate $\eta$ is $0.3$, the number of iterations $K=10$, and the sampling times $M$ in \Cref{eq:daytime-optimize-loss} is $8$. For performance evaluation, we employ \textit{Pass@1} accuracy under a sampling temperature
of $0$ across all benchmarks. 

\vspace{-0.7em}
\paragraph{Baselines.} We compare against several well-established baselines:
\vspace{-1em}
\begin{itemize}[leftmargin=1em,itemsep=-0.2em]
\item \textbf{Prompting (training-free)}: vanilla model and  CoT~\citep{wei2023cot};
\item \textbf{Reinforcement Learning}: (1) self-rewarding methods, including Self-Rewarding~\citep{yuan2025selfrewardinglanguagemodels} and Genius~\citep{xu-etal-2025-genius}, and (2) verifiable reward methods, including GRPO~\citep{deepseekai2025deepseekr1incentivizingreasoningcapability}, Reinforce~\citep{williams1992simple}, and Reinforce++~\citep{hu2025reinforceefficientrlhfalgorithm}. The latter three baselines are trained independently on the training split of each dataset and evaluated on the corresponding test split. Owing to the limited size of AIME24/25, models trained on MATH are directly evaluated on these benchmarks. Results on SciBench are omitted for these baselines due to the absence of a dedicated training set.
\item \textbf{Latent Reasoning}, including Coprocessor~\citep{liu2024deliberation} and SoftCoT~\citep{xu2025softcottesttimescalingsoft}.
\item \textbf{Test-time Scaling} methods, including Self-Consistency~\citep{wang2023selfconsistencyimproveschainthought}, Self-refine~\citep{madaan2023selfrefineiterativerefinementselffeedback}, LatentSeek~\citep{li2025seekdarkreasoningtesttime}, and TTRL~\citep{xiang20252reasoningllmslearning}.
\end{itemize}

\begin{table*}[!t]
\centering
\caption{\textbf{Performance Comparison} across two LLM backbones (\llmname{Qwen2.5-7b} and \llmname{Llama3.2-3b}), against thirteen baselines and on eight benchmarks.  The best and second best results are \colorbox{cyan!20}{highlighted} and \underline{underlined}, respectively.}
\vspace{-0.8em}
\label{tab:main_results}
\resizebox{\textwidth}{!}{%
\begin{tabular}{@{}c|lccccccccc@{}}
\toprule
 &{\multirow{2}{*}{\textbf{Method}}} & \textbf{General QA} & \multicolumn{4}{c}{\textbf{Mathematical Reasoning}} & \multicolumn{2}{c}{\textbf{Sci. Reasoning}} & \textbf{Med. Reasoning} \\
\cmidrule(lr){3-3} \cmidrule(lr){4-7} \cmidrule(lr){8-9} \cmidrule(lr){10-10}
&  & MMLU & GSM8K & MATH-500 &  AIME24 & AIME25 & SciBench & GPQA & JAMA Clinical \\
\midrule

\multirow{18}{*}{\rotatebox{90}{\llmname{Qwen2.5-7b}}} &
 \multicolumn{9}{c}{\textbf{\textit{Prompting (training-free})}} \\
 & Vanilla Model & 55.30 & 87.72 & 55.80 & 0.00 & 0.00 & 11.27 & 27.78 & 47.72 \\
 & CoT & 69.10 & 87.04 & 68.80 & 6.67 & 3.33 & 11.99 & 30.81 & 50.96 \\
\cmidrule{2-11}
& \multicolumn{9}{c}{\textbf{\textit{Reinforcement Learning}}} \\
 & Self-Rewarding & 63.10 & 88.30 & 59.62 & 0.00 & 0.00 & 9.36 & 23.65 & 47.07 \\
 & Genius & 58.30 & 87.93 & 49.57 & 0.00 & 0.00 & 13.60 & 29.31 & 41.78 \\
 & GRPO & 68.90 & 92.30 & 75.85 & 6.67 & 3.33 & - & 33.60 & \underline{51.62} \\
 & Reinforce & 63.77 & 92.30 & 76.80 & 6.67 & 6.67 & - & \underline{34.34} & 49.16 \\
  & Reinforce++ & 65.90 & 92.60 & 75.02 & \underline{13.33} & 6.67 & - & \underline{34.34} & 50.40 \\
\cmidrule{2-11}
& \multicolumn{9}{c}{\textbf{\textit{Latent Reasoning}}} \\
 & Coprocessor & 68.10 & 83.60 & 53.73 & 6.67 & 6.67 & - & 31.88 & 43.70 \\
 & SoftCoT & 62.30 & 80.13 & 65.80 & 3.33 & 0.00 & - & 28.28 & 49.70 \\
\cmidrule{2-11}
& \multicolumn{9}{c}{\textbf{\textit{Test-time Scaling}}} \\
 & Self-Consistency & 69.80 & 88.62 & 69.40 & 6.67 & 6.67 & 12.13 & 32.32 & \underline{51.62} \\
 & Self-Refine & 61.40 & 86.33 & 59.32 & 3.33 & 0.00 & 9.36 & 22.65 & 45.64 \\
 & LatentSeek & 68.50 & 91.58 & 66.20 & 10.00 & 3.33 & \underline{14.45} & 31.31 & 50.40 \\
 & TTRL & \underline{70.90}  & \underline{92.80} & \underline{77.39} & \cellcolor{cyan!20}\textbf{23.33} & \cellcolor{cyan!20}\textbf{13.33}  & 13.92 & 33.60 & 49.16 \\
& {\ourmethod} & \cellcolor{cyan!20}\textbf{72.30} & \cellcolor{cyan!20}\textbf{92.98} & \cellcolor{cyan!20}\textbf{77.60} & \cellcolor{cyan!20}\textbf{23.33} & \underline{10.00} & \cellcolor{cyan!20}\textbf{19.79} & \cellcolor{cyan!20}\textbf{34.85} & \cellcolor{cyan!20}\textbf{52.94} \\
\midrule
\midrule

\multirow{18}{*}{\rotatebox{90}{\llmname{Llama3.2-3b}}} &
 \multicolumn{9}{c}{\textbf{\textit{Prompting (training-free})}} \\
 & Vanilla Model & 60.60 & 71.65 & 41.60 & 0.00 & 0.00 & 6.79 & 26.77 & 45.14 \\
 & CoT & 57.60 & 64.90 & 48.60 & 0.00 & 0.00 & 7.95 & 26.77 &45.60 \\
\cmidrule{2-11}
& \multicolumn{9}{c}{\textbf{\textit{Reinforcement Learning}}} \\
 & Self-Rewarding &  57.30  & 69.22 & 39.20 & 0.00 & 0.00 & 3.19 & 23.90 & 40.16 \\
 & Genius & 58.20 & 73.61 & 38.15 & 0.00 & 0.00 & 6.79  &  21.80 & 45.60\\
 & GRPO & \underline{62.70} & \underline{75.30} & 50.20 & \underline{3.33} & 0.00 & - & \underline{28.18} & \underline{46.26} \\
 & Reinforce & 60.60 & 75.02 & 49.60 & \underline{3.33} & 0.00 & - & 24.50 & 45.60 \\
  & Reinforce++ & \underline{62.70} & 73.61 & 50.20 & \underline{3.33} & \underline{3.33} & - & 26.26 & 44.80 \\
\cmidrule{2-11}
& \multicolumn{9}{c}{\textbf{\textit{Latent Reasoning}}} \\
 & Coprocessor & 61.50 &  70.08 & 44.90 & 0.00 & 0.00 & - & 21.80 & 42.28 \\
 & SoftCoT & 58.90 & 73.61 & 46.40 & 0.00 & 0.00 & - & 25.25 & 43.35 \\
\cmidrule{2-11}
& \multicolumn{9}{c}{\textbf{\textit{Test-time Scaling}}} \\
 & Self-Consistency & 59.10 & 66.33 & 49.20 & 0.00 & 0.00 & \underline{8.67} & 27.27 & 45.60 \\
 & Self-Refine & 58.90 & 68.90 & 44.10 & 0.00 & 0.00 & 4.28 & 20.10 & 42.28  \\
 & LatentSeek & 49.30 & 55.95 & 38.60 & 0.00 & 0.00 & 5.20 & 26.26 & 32.36 \\
 & TTRL & 62.10 & 75.02 & \underline{51.00} & \underline{3.33} & \cellcolor{cyan!20}\textbf{6.67} & 8.07 & \underline{28.18} & 44.80 \\
& {\ourmethod} & \cellcolor{cyan!20}\textbf{64.30} & \cellcolor{cyan!20}\textbf{75.51} & \cellcolor{cyan!20}\textbf{51.20}& \cellcolor{cyan!20}\textbf{6.67} & \underline{3.33} & \cellcolor{cyan!20}\textbf{9.39} & \cellcolor{cyan!20}\textbf{29.29} & \cellcolor{cyan!20}\textbf{48.44} \\
\bottomrule
\end{tabular}%
}
\vspace{-1em}
\end{table*}

\vspace{-0.5em}
\subsection{Main Results}
\vspace{-0.5em}
\paragraph{Obs. \ding{182}: \ourmethod performs well across most task domains.} As shown in \Cref{tab:main_results}, most baselines fail to deliver consistent gains across all benchmark types. LatentSeek and TTRL excel in mathematical reasoning yet fall short in other domains: for instance, LatentSeek with \llmname{Llama3.2-3b} underperforms the vanilla model on MMLU ($-11.3\%$) and experiences a performance drop on SciBench ($-1.59\%$), while TTRL with \llmname{Qwen2.5-7b} yields limited benefit on JAMA Clinical ($+1.44\%$). In contrast, \ourmethod not only matches or surpasses TTRL in the general QA domain (e.g., on \llmname{Qwen2.5-7b}, MMLU $+6.4\%$) but also achieves superior results in other domains, such as a $+8.52\%$ improvement on SciBench+\llmname{Qwen2.5-7b}.

\vspace{-0.4em}
\paragraph{Obs. \ding{183}: \ourmethod generalizes well across LLM backbones.} In contrast to many baselines whose gains across different LLMs are highly inconsistent (\textit{e.g.}, Coprocessor on GPQA yields $+4.1\%$ with \llmname{Qwen2.5-7b} but $-4.97\%$ with \llmname{Llama3.2-3b}), \ourmethod consistently delivers positive improvements across models of varying scales, as clearly illustrated in \Cref{tab:multi-llm}. Notably, its benefits naturally \textit{scale} with model size: on MATH-500, for example, the improvement rises from $9.6\%$ with \llmname{Llma3.2-3b} to $20.8\%$ with \llmname{Gemma-3-12b}.

\vspace{-0.4em}
\subsection{Generalization and Continual Learning Study}\label{sec:generalization-exp}
\begin{figure}
\includegraphics[width=\textwidth]{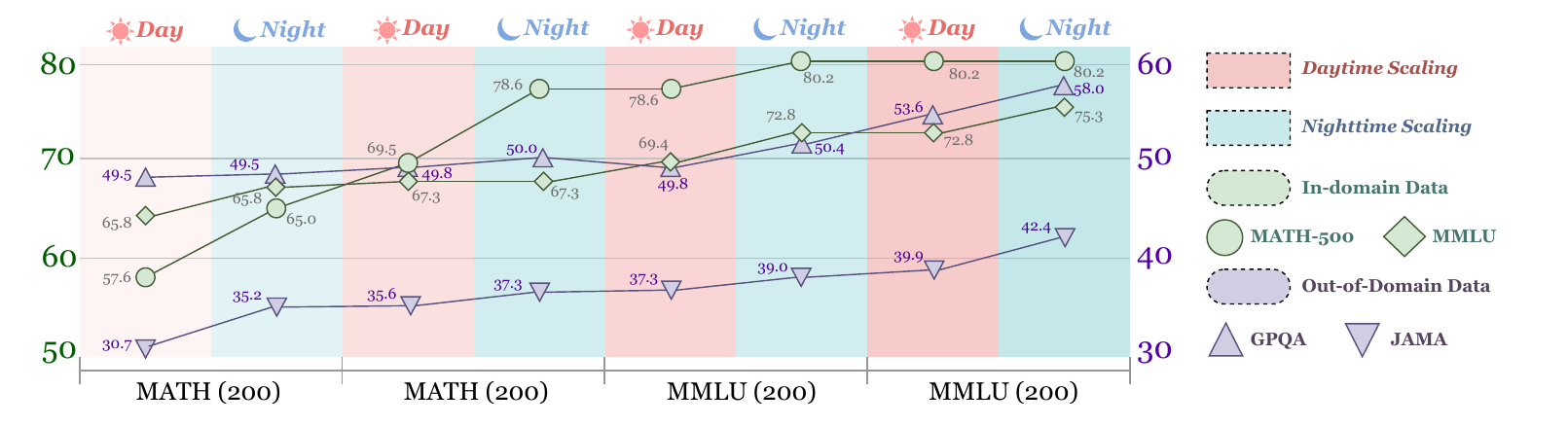}
\vspace{-1.8em}
\caption{The evolving dynamics of \ourmethod when applied on \llmname{Gemma-3-12b} across two in-domain and out-of-domain datasets.} \label{fig:evolve-dynamic}
\vspace{-1.5em}
\end{figure}

\vspace{-0.5em}
This section investigates the continual learning and generalization capacity of \ourmethod. \Cref{fig:evolve-dynamic} illustrates performance trajectories when \ourmethod, instantiated with \llmname{Gemma-3-12b}, sequentially processes test data from MATH and MMLU. In-domain evaluation is conducted on MATH and MMLU, while out-of-domain evaluation is performed on GPQA and JAMA. Red zones denote evaluations after a daytime scaling step with updated episodic buffer $\mathcal{M}'$, whereas blue zones correspond to evaluations after a nighttime scaling step yielding updated latent weaver $\mathbf{W}'_\psi$.

\vspace{-0.4em}
\paragraph{Obs. \ding{184}: \ourmethod generalizes across domains.} As shown in \Cref{fig:evolve-dynamic}, after two rounds of MATH data, performance improves substantially in-domain ($57.6\% \rightarrow 78.6\%$) while also transferring gains to distinct domains (JAMA $+6.6\%$, MMLU $+1.5\%$). Notably, nighttime scaling proves more conducive to such cross-domain generalization: the first nighttime scaling on MATH increases JAMA by $+4.5\%$, compared to only $+0.4\%$ from daytime scaling. This highlights that nighttime scaling, akin to cortical consolidation in the human brain, integrates experiences into more transferable knowledge, whereas daytime scaling yields more superficial improvements. Moreover, \ourmethod demonstrates strong continual learning ability: after two rounds of MMLU data, \llmname{Gemma-3-12b} shows not only no degradation but a slight improvement on MATH ($78.6\%\rightarrow 80.2\%$), highlighting the robustness of \ourmethod in continual adaptation.

\begin{table*}[!t]
\centering
\renewcommand{\tabcolsep}{2pt}
\scriptsize
\caption{\textbf{Performance Comparison} of the vanilla model versus that enhanced with \ourmethod across five LLM backbones. The $\Delta$ row indicates the absolute improvement.}
\vspace{-0.5em}
\label{tab:multi-llm}
\begin{tabular}{@{}llcccccccc@{}}
\toprule
\textbf{LLM Backbone} & \textbf{Method} & \textbf{MMLU} & \textbf{GSM8K} & \textbf{MATH-500}  & \textbf{SciBench} & \textbf{GPQA} & \textbf{JAMA Clinical} & \textbf{AIME24} & \textbf{AIME25}  \\
\midrule

\multirow{3}{*}{\llmname{Llama3.2-3b}}
 & Vanilla & 60.60 & 71.65 & 41.60 & 6.79 & 26.77 & 45.14 & 0.00 & 0.00 \\
 & +\ourmethod & 64.30 & 75.51 & 51.20 & 9.39 & 29.29 & 48.44 & 6.67 & 3.33 \\
 & $\Delta$ & \textcolor{orange}{+3.70} & \textcolor{orange}{+3.86} & \textcolor{orange}{+9.60} & \textcolor{orange}{+2.60} & \textcolor{orange}{+2.52} & \textcolor{orange}{+3.30} & \textcolor{orange}{+6.67} & \textcolor{orange}{+3.33} \\
\midrule

\multirow{3}{*}{\llmname{Qwen2.5-7b}} 
 & Vanilla & 55.30 & 87.72 & 55.80 & 11.27 & 27.78 & 47.72 & 0.00 & 0.00 \\
 & +\ourmethod & 72.30 & 92.98 & 77.60 & 19.79 & 34.85 & 52.94 & 23.33 & 10.00 \\
  & $\Delta$ & \textcolor{orange}{+17.00} & \textcolor{orange}{+5.26} & \textcolor{orange}{+21.80} & \textcolor{orange}{+8.52} & \textcolor{orange}{+7.07} & \textcolor{orange}{+5.22} & \textcolor{orange}{+23.33} & \textcolor{orange}{+10.00} \\
\midrule


 \multirow{3}{*}{\llmname{Qwen3-4b}}
 & Vanilla & 71.90 & 89.23 & 61.40 & 12.28 & 34.85 & 51.49 & 10.00 & 3.33 \\
 & +\ourmethod & 73.30 & 92.42 & 78.60 & 31.93 & 38.89 & 53.67 & 23.33 & 16.67 \\
 & $\Delta$ & \textcolor{orange}{+1.40} & \textcolor{orange}{+3.19} & \textcolor{orange}{+17.20} & \textcolor{orange}{+19.65} & \textcolor{orange}{+4.04} & \textcolor{orange}{+2.18} & \textcolor{orange}{+13.33} & \textcolor{orange}{+13.34}  \\
\midrule

 \multirow{3}{*}{\llmname{Qwen3-8b}}
 & Vanilla & 72.70 & 87.94 & 55.20 & 6.36 & 28.82 & 53.08 & 3.33 & 3.33 \\
 & +\ourmethod & 78.80 & 90.45 & 73.80 & 10.83 & 32.82 & 54.60 & 26.67 & 23.33 \\
 & $\Delta$ & \textcolor{orange}{+6.10} & \textcolor{orange}{+2.51} & \textcolor{orange}{+18.60} & \textcolor{orange}{+4.47} & \textcolor{orange}{+4.00} & \textcolor{orange}{+1.52} & \textcolor{orange}{+23.33} & \textcolor{orange}{+20.00} \\
\midrule

 \multirow{3}{*}{\llmname{Gemma-3-12b}}
 & Vanilla & 65.80 & 89.23 & 57.40 & 10.84 & 33.33 & 49.50 & 0.00 & 10.00 \\
 & +\ourmethod & 73.90 & 91.89 & 78.20 & 18.93 & 41.92 & 55.06 & 10.00 & 13.33 \\
 & $\Delta$ & \textcolor{orange}{+8.10} & \textcolor{orange}{+2.66} & \textcolor{orange}{+20.80} & \textcolor{orange}{+8.09} & \textcolor{orange}{+8.59} & \textcolor{orange}{+5.56} & \textcolor{orange}{+10.00} & \textcolor{orange}{+3.33} \\

\bottomrule
\end{tabular}
\end{table*}

\subsection{Framework Analyasis}
\begin{figure}
\includegraphics[width=\textwidth]{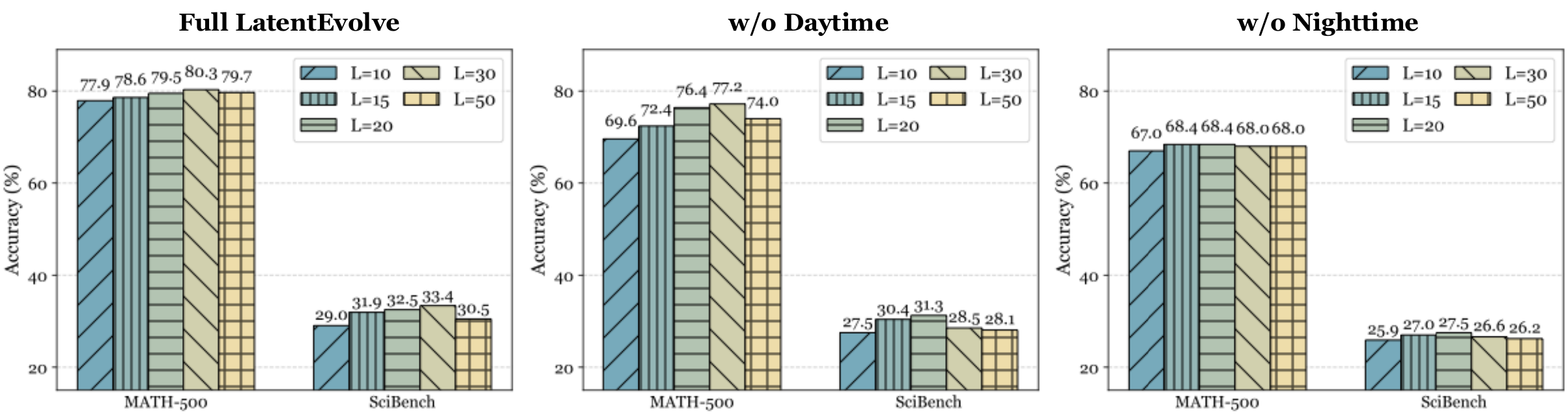}
\vspace{-1.8em}
\caption{The ablation study and parameter sensitivity analysis of \ourmethod.} \label{fig:ablation}
\vspace{-1.5em}
\end{figure}

\vspace{-0.4em}
\paragraph{Sensitivity Analysis.} We conduct a sensitivity analysis of \textbf{the parameter $L'$}, which determines the dimensionality of each initialized latent representation $\mathbf{z}_{\text{base},i}$. As shown in \Cref{fig:ablation} (\textit{Left}), both MATH and SciBench exhibit similar patterns: as $L'$ increases from $10$ to $50$, performance first improves and then declines, with the best results attained at $L'=30$ (MATH $80.3\%$, SciBench $33.4\%$). A plausible explanation is that too few dimensions cannot adequately encode historical optimization experience, while excessively many dimensions introduce additional parameters that may hinder the self-supervised refinement process during daytime scaling (\Cref{eq:daytime-optimize-loss}). Analysis with more parameters (\textit{e.g.}, \textbf{the evolution period $T$}) is provided in \Cref{app:sensi}.

\vspace{-0.4em}
\paragraph{Ablation Study.} We investigate two variants of \ourmethod: \textit{w/o Daytime} and \textit{w/o Nighttime}. The former removes \Cref{eq:momentum}, meaning that each latent optimization proceeds without retrieving prior experience, and the latter omits the update of the latent weaver as described in \Cref{eq:consolidation_loss}. As shown in \Cref{fig:ablation}, removing either daytime or nighttime scaling leads to a substantial performance drop compared with the full \ourmethod, with the effect of \textit{w/o Nighttime} being more pronounced. At $L'=30$, \textit{w/o Daytime} reduces performance on SciBench from $33.4\%$ to $28.5\%$ ($-4.9\%$), while \textit{w/o Nighttime} results in a larger decline of $6.8\%$. This observation aligns with the findings in \Cref{sec:generalization-exp}. Overall, both components of \ourmethod are indispensable, mutually reinforcing, and jointly critical to its effectiveness.

\begin{table*}[htbp]
\centering
\scriptsize
\ttfamily
\caption{Case Studies of \ourmethod. The questions (from SciBench and GPQA), the answers, and the \ourmethod generations are listed. The peculiar reasoning tokens are highlighted in blue.}
\vspace{-0.4em}
\label{tab:case_studies_latentevolve_special}
\begin{tabularx}{\textwidth}{X} 
\toprule
\textbf{\color{QuestionColor}QUESTION:} Express the van der Waals parameters b=0.0226 dm$^3$ mol$^{-1}$ in SI base units. \\
\textbf{\color{AnswerColor}ANSWER:} 0.0000226 \\
\textbf{LatentEvlove:} To convert parameter b=0.0226 dm$^3$ mol$^{-1}$ in SI base units. Start \textcolor{ReasoningColor}{unit-conv dm3-to-m3 step one}. Identify 1 dm equal 0.1 m. So, 1 dm$^3$ is 0.001 m$^3$. Then \textcolor{ReasoningColor}{apply factor-multiplied 0.0226}. [$\ldots$] The final value in SI base units is 2.26 * 10$^{-5}$ m$^3$ mol$^{-1}$.\\
\midrule
\textbf{\color{QuestionColor}QUESTION:} Which of the following physical theories never requires regularization at high energies? A. Quantum Chromodynamics B. Superstring Theory C. Quantum Electrodynamics D. Classical Electrodynamics\\
\textbf{\color{AnswerColor}ANSWER:} B \\
\textbf{LatentEvlove:} Which physical theories \textcolor{ReasoningColor}{never never} required regularization at high energies. \textcolor{ReasoningColor}{Need-no-regularization-scan start}. QCD, QED, needs. Classical Electrodynamics, not quantum, so no need. Superstring Theory, this theory \textcolor{ReasoningColor}{no require regularization itself}. [$\ldots$] Based on the above analysis, the theory that never requires regularization at high energies is Superstring Theory. The correct answer is B.  \\
\bottomrule
\end{tabularx}
\end{table*}

\vspace{-0.5em}
\paragraph{Case Study.} To gain insight into \ourmethod's latent optimization, we qualitatively analyzed its outputs following \citep{li2025seekdarkreasoningtesttime}. As shown in \Cref{tab:case_studies_latentevolve_special}, we observed distinctive reasoning tokens and patterns, including fragmented internal commands (\textcolor{ReasoningColor}{\texttt{Need-no-regularization-scan start}}), lexical repetition (\textcolor{ReasoningColor}{\texttt{never never}}), and unconventional grammar (\textcolor{ReasoningColor}{\texttt{no require regularization itself}}). Despite these peculiar tokens, \ourmethod consistently produces correct answers, suggesting it steers the LLM along more machine-native, efficient reasoning trajectories within latent space. \Cref{tab:algorithmic_statistics_new_llms} further shows that, relative to vanilla CoT, \ourmethod concludes the reasoning with fewer decoding tokens.

\section{Conclusion}

\vspace{-0.5em}
In this work, we proposed \ourmethod, a self-evolving latent test-time scaling framework inspired by complementary learning systems. By alternating \textit{daytime scaling} for fast episodic adaptation with \textit{nighttime scaling} for slow procedural consolidation, our approach enables LLMs to accumulate and refine experiential knowledge during inference without external supervision. Experiments across eight benchmarks and five model backbones show that \ourmethod surpasses state-of-the-art TTS methods (\textit{e.g.}, TTRL, LatentSeek), transfers effectively across tasks, and exhibits steady continual learning ability. Broadly, our work points toward a new path where LLMs not only \textit{scale at test time}, but also \textit{evolve through it}, bringing them closer to the adaptive and accumulative intelligence seen in human cognition.

\bibliography{iclr2025_conference}

\begin{thebibliography}{92}
\providecommand{\natexlab}[1]{#1}
\providecommand{\url}[1]{\texttt{#1}}
\expandafter\ifx\csname urlstyle\endcsname\relax
  \providecommand{\doi}[1]{doi: #1}\else
  \providecommand{\doi}{doi: \begingroup \urlstyle{rm}\Url}\fi

\bibitem[Aghajanyan et~al.(2023)Aghajanyan, Yu, Conneau, Hsu, Hambardzumyan, Zhang, Roller, Goyal, Levy, and Zettlemoyer]{aghajanyan2023scalinglawsgenerativemixedmodal}
Armen Aghajanyan, Lili Yu, Alexis Conneau, Wei-Ning Hsu, Karen Hambardzumyan, Susan Zhang, Stephen Roller, Naman Goyal, Omer Levy, and Luke Zettlemoyer.
\newblock Scaling laws for generative mixed-modal language models, 2023.
\newblock URL \url{https://arxiv.org/abs/2301.03728}.

\bibitem[Aky{\"u}rek et~al.(2024)Aky{\"u}rek, Damani, Zweiger, Qiu, Guo, Pari, Kim, and Andreas]{akyurek2024surprisingttt}
Ekin Aky{\"u}rek, Mehul Damani, Adam Zweiger, Linlu Qiu, Han Guo, Jyothish Pari, Yoon Kim, and Jacob Andreas.
\newblock The surprising effectiveness of test-time training for few-shot learning.
\newblock \emph{arXiv preprint arXiv:2411.07279}, 2024.

\bibitem[ang Gao et~al.(2025)ang Gao, Geng, Hua, Hu, Juan, Liu, Liu, Qiu, Qi, Wu, Wang, Xiao, Zhou, Zhang, Zhang, Xiang, Fang, Zhao, Liu, Ren, Qian, Wang, Hu, Wang, Wu, Ji, and Wang]{gao2025surveyselfevolvingagentspath}
Huan ang Gao, Jiayi Geng, Wenyue Hua, Mengkang Hu, Xinzhe Juan, Hongzhang Liu, Shilong Liu, Jiahao Qiu, Xuan Qi, Yiran Wu, Hongru Wang, Han Xiao, Yuhang Zhou, Shaokun Zhang, Jiayi Zhang, Jinyu Xiang, Yixiong Fang, Qiwen Zhao, Dongrui Liu, Qihan Ren, Cheng Qian, Zhenhailong Wang, Minda Hu, Huazheng Wang, Qingyun Wu, Heng Ji, and Mengdi Wang.
\newblock A survey of self-evolving agents: On path to artificial super intelligence, 2025.
\newblock URL \url{https://arxiv.org/abs/2507.21046}.

\bibitem[Besta et~al.(2024)Besta, Blach, Kubicek, Gerstenberger, Podstawski, Gianinazzi, Gajda, Lehmann, Niewiadomski, Nyczyk, and Hoefler]{graphofthoughts}
Maciej Besta, Nils Blach, Ales Kubicek, Robert Gerstenberger, Michal Podstawski, Lukas Gianinazzi, Joanna Gajda, Tomasz Lehmann, Hubert Niewiadomski, Piotr Nyczyk, and Torsten Hoefler.
\newblock Graph of thoughts: Solving elaborate problems with large language models.
\newblock \emph{Proceedings of the AAAI Conference on Artificial Intelligence}, 38\penalty0 (16):\penalty0 17682–17690, March 2024.
\newblock ISSN 2159-5399.
\newblock \doi{10.1609/aaai.v38i16.29720}.
\newblock URL \url{http://dx.doi.org/10.1609/aaai.v38i16.29720}.

\bibitem[Brown et~al.(2024)Brown, Juravsky, Ehrlich, Clark, Le, Ré, and Mirhoseini]{brown2024largelanguagemonkeysscaling}
Bradley Brown, Jordan Juravsky, Ryan Ehrlich, Ronald Clark, Quoc~V. Le, Christopher Ré, and Azalia Mirhoseini.
\newblock Large language monkeys: Scaling inference compute with repeated sampling, 2024.
\newblock URL \url{https://arxiv.org/abs/2407.21787}.

\bibitem[Bubeck et~al.(2023)Bubeck, Chandrasekaran, Eldan, Gehrke, Horvitz, Kamar, Lee, Lee, Li, Lundberg, et~al.]{bubeck2023sparks}
S{\'e}bastien Bubeck, Varun Chandrasekaran, Ronen Eldan, Johannes Gehrke, Eric Horvitz, Ece Kamar, Peter Lee, Yin~Tat Lee, Yuanzhi Li, Scott Lundberg, et~al.
\newblock Sparks of artificial general intelligence: Early experiments with gpt-4.
\newblock \emph{arXiv preprint arXiv:2303.12712}, 2023.

\bibitem[Chen et~al.(2025{\natexlab{a}})Chen, Fang, Singla, and Dredze]{chen2025benchmarkingjama}
Hanjie Chen, Zhouxiang Fang, Yash Singla, and Mark Dredze.
\newblock Benchmarking large language models on answering and explaining challenging medical questions.
\newblock In \emph{Proceedings of the 2025 Conference of the Nations of the Americas Chapter of the Association for Computational Linguistics: Human Language Technologies (Volume 1: Long Papers)}, pp.\  3563--3599, 2025{\natexlab{a}}.

\bibitem[Chen et~al.(2025{\natexlab{b}})Chen, Lu, Kim, Zhang, Tang, Piché, Gontier, Bengio, and Kamalloo]{chen2025selfevolvingcurriculumllmreasoning}
Xiaoyin Chen, Jiarui Lu, Minsu Kim, Dinghuai Zhang, Jian Tang, Alexandre Piché, Nicolas Gontier, Yoshua Bengio, and Ehsan Kamalloo.
\newblock Self-evolving curriculum for llm reasoning, 2025{\natexlab{b}}.
\newblock URL \url{https://arxiv.org/abs/2505.14970}.

\bibitem[Chen et~al.(2024)Chen, Liu, Wang, Zhang, Liu, Lin, Chen, and Zhao]{chen2024agentflandesigningdatamethods}
Zehui Chen, Kuikun Liu, Qiuchen Wang, Wenwei Zhang, Jiangning Liu, Dahua Lin, Kai Chen, and Feng Zhao.
\newblock Agent-flan: Designing data and methods of effective agent tuning for large language models, 2024.
\newblock URL \url{https://arxiv.org/abs/2403.12881}.

\bibitem[Choi et~al.(2023)Choi, Kim, Park, Mok, and Lee]{choi-etal-2023-smop}
Joon-Young Choi, Junho Kim, Jun-Hyung Park, Wing-Lam Mok, and SangKeun Lee.
\newblock {SM}o{P}: Towards efficient and effective prompt tuning with sparse mixture-of-prompts.
\newblock In Houda Bouamor, Juan Pino, and Kalika Bali (eds.), \emph{Proceedings of the 2023 Conference on Empirical Methods in Natural Language Processing}, pp.\  14306--14316, Singapore, December 2023. Association for Computational Linguistics.
\newblock \doi{10.18653/v1/2023.emnlp-main.884}.
\newblock URL \url{https://aclanthology.org/2023.emnlp-main.884/}.

\bibitem[Chung et~al.(2025)Chung, Hsiao, Huang, Cho, Lin, Ziwei, and Chen]{chung2025revisitingtesttimescalingsurvey}
Ho-Lam Chung, Teng-Yun Hsiao, Hsiao-Ying Huang, Chunerh Cho, Jian-Ren Lin, Zhang Ziwei, and Yun-Nung Chen.
\newblock Revisiting test-time scaling: A survey and a diversity-aware method for efficient reasoning, 2025.
\newblock URL \url{https://arxiv.org/abs/2506.04611}.

\bibitem[Cobbe et~al.(2021)Cobbe, Kosaraju, Bavarian, Chen, Jun, Kaiser, Plappert, Tworek, Hilton, Nakano, et~al.]{cobbe2021trainingverifier}
Karl Cobbe, Vineet Kosaraju, Mohammad Bavarian, Mark Chen, Heewoo Jun, Lukasz Kaiser, Matthias Plappert, Jerry Tworek, Jacob Hilton, Reiichiro Nakano, et~al.
\newblock Training verifiers to solve math word problems.
\newblock \emph{arXiv preprint arXiv:2110.14168}, 2021.

\bibitem[DeepSeek-AI et~al.(2025)DeepSeek-AI, Guo, Yang, Zhang, Song, Zhang, Xu, Zhu, Ma, Wang, Bi, Zhang, Yu, Wu, Wu, Gou, Shao, Li, Gao, Liu, Xue, Wang, Wu, Feng, Lu, Zhao, Deng, Zhang, Ruan, Dai, Chen, Ji, Li, Lin, Dai, Luo, Hao, Chen, Li, Zhang, Bao, Xu, Wang, Ding, Xin, Gao, Qu, Li, Guo, Li, Wang, Chen, Yuan, Qiu, Li, Cai, Ni, Liang, Chen, Dong, Hu, Gao, Guan, Huang, Yu, Wang, Zhang, Zhao, Wang, Zhang, Xu, Xia, Zhang, Zhang, Tang, Li, Wang, Li, Tian, Huang, Zhang, Wang, Chen, Du, Ge, Zhang, Pan, Wang, Chen, Jin, Chen, Lu, Zhou, Chen, Ye, Wang, Yu, Zhou, Pan, Li, Zhou, Wu, Ye, Yun, Pei, Sun, Wang, Zeng, Zhao, Liu, Liang, Gao, Yu, Zhang, Xiao, An, Liu, Wang, Chen, Nie, Cheng, Liu, Xie, Liu, Yang, Li, Su, Lin, Li, Jin, Shen, Chen, Sun, Wang, Song, Zhou, Wang, Shan, Li, Wang, Wei, Zhang, Xu, Li, Zhao, Sun, Wang, Yu, Zhang, Shi, Xiong, He, Piao, Wang, Tan, Ma, Liu, Guo, Ou, Wang, Gong, Zou, He, Xiong, Luo, You, Liu, Zhou, Zhu, Xu, Huang, Li, Zheng, Zhu, Ma, Tang, Zha, Yan, Ren, Ren, Sha, Fu, Xu, Xie, Zhang,
  Hao, Ma, Yan, Wu, Gu, Zhu, Liu, Li, Xie, Song, Pan, Huang, Xu, Zhang, and Zhang]{deepseekai2025deepseekr1incentivizingreasoningcapability}
DeepSeek-AI, Daya Guo, Dejian Yang, Haowei Zhang, Junxiao Song, Ruoyu Zhang, Runxin Xu, Qihao Zhu, Shirong Ma, Peiyi Wang, Xiao Bi, Xiaokang Zhang, Xingkai Yu, Yu~Wu, Z.~F. Wu, Zhibin Gou, Zhihong Shao, Zhuoshu Li, Ziyi Gao, Aixin Liu, Bing Xue, Bingxuan Wang, Bochao Wu, Bei Feng, Chengda Lu, Chenggang Zhao, Chengqi Deng, Chenyu Zhang, Chong Ruan, Damai Dai, Deli Chen, Dongjie Ji, Erhang Li, Fangyun Lin, Fucong Dai, Fuli Luo, Guangbo Hao, Guanting Chen, Guowei Li, H.~Zhang, Han Bao, Hanwei Xu, Haocheng Wang, Honghui Ding, Huajian Xin, Huazuo Gao, Hui Qu, Hui Li, Jianzhong Guo, Jiashi Li, Jiawei Wang, Jingchang Chen, Jingyang Yuan, Junjie Qiu, Junlong Li, J.~L. Cai, Jiaqi Ni, Jian Liang, Jin Chen, Kai Dong, Kai Hu, Kaige Gao, Kang Guan, Kexin Huang, Kuai Yu, Lean Wang, Lecong Zhang, Liang Zhao, Litong Wang, Liyue Zhang, Lei Xu, Leyi Xia, Mingchuan Zhang, Minghua Zhang, Minghui Tang, Meng Li, Miaojun Wang, Mingming Li, Ning Tian, Panpan Huang, Peng Zhang, Qiancheng Wang, Qinyu Chen, Qiushi Du, Ruiqi Ge, Ruisong
  Zhang, Ruizhe Pan, Runji Wang, R.~J. Chen, R.~L. Jin, Ruyi Chen, Shanghao Lu, Shangyan Zhou, Shanhuang Chen, Shengfeng Ye, Shiyu Wang, Shuiping Yu, Shunfeng Zhou, Shuting Pan, S.~S. Li, Shuang Zhou, Shaoqing Wu, Shengfeng Ye, Tao Yun, Tian Pei, Tianyu Sun, T.~Wang, Wangding Zeng, Wanjia Zhao, Wen Liu, Wenfeng Liang, Wenjun Gao, Wenqin Yu, Wentao Zhang, W.~L. Xiao, Wei An, Xiaodong Liu, Xiaohan Wang, Xiaokang Chen, Xiaotao Nie, Xin Cheng, Xin Liu, Xin Xie, Xingchao Liu, Xinyu Yang, Xinyuan Li, Xuecheng Su, Xuheng Lin, X.~Q. Li, Xiangyue Jin, Xiaojin Shen, Xiaosha Chen, Xiaowen Sun, Xiaoxiang Wang, Xinnan Song, Xinyi Zhou, Xianzu Wang, Xinxia Shan, Y.~K. Li, Y.~Q. Wang, Y.~X. Wei, Yang Zhang, Yanhong Xu, Yao Li, Yao Zhao, Yaofeng Sun, Yaohui Wang, Yi~Yu, Yichao Zhang, Yifan Shi, Yiliang Xiong, Ying He, Yishi Piao, Yisong Wang, Yixuan Tan, Yiyang Ma, Yiyuan Liu, Yongqiang Guo, Yuan Ou, Yuduan Wang, Yue Gong, Yuheng Zou, Yujia He, Yunfan Xiong, Yuxiang Luo, Yuxiang You, Yuxuan Liu, Yuyang Zhou, Y.~X. Zhu,
  Yanhong Xu, Yanping Huang, Yaohui Li, Yi~Zheng, Yuchen Zhu, Yunxian Ma, Ying Tang, Yukun Zha, Yuting Yan, Z.~Z. Ren, Zehui Ren, Zhangli Sha, Zhe Fu, Zhean Xu, Zhenda Xie, Zhengyan Zhang, Zhewen Hao, Zhicheng Ma, Zhigang Yan, Zhiyu Wu, Zihui Gu, Zijia Zhu, Zijun Liu, Zilin Li, Ziwei Xie, Ziyang Song, Zizheng Pan, Zhen Huang, Zhipeng Xu, Zhongyu Zhang, and Zhen Zhang.
\newblock Deepseek-r1: Incentivizing reasoning capability in llms via reinforcement learning, 2025.
\newblock URL \url{https://arxiv.org/abs/2501.12948}.

\bibitem[Fang et~al.(2025)Fang, Peng, Zhang, Wang, Yi, Zhang, Xu, Wu, Liu, Li, Ren, Aletras, Wang, Zhou, and Meng]{fang2025comprehensivesurveyselfevolvingai}
Jinyuan Fang, Yanwen Peng, Xi~Zhang, Yingxu Wang, Xinhao Yi, Guibin Zhang, Yi~Xu, Bin Wu, Siwei Liu, Zihao Li, Zhaochun Ren, Nikos Aletras, Xi~Wang, Han Zhou, and Zaiqiao Meng.
\newblock A comprehensive survey of self-evolving ai agents: A new paradigm bridging foundation models and lifelong agentic systems, 2025.
\newblock URL \url{https://arxiv.org/abs/2508.07407}.

\bibitem[Geiping et~al.(2025)Geiping, McLeish, Jain, Kirchenbauer, Singh, Bartoldson, Kailkhura, Bhatele, and Goldstein]{geiping2025scalingtesttimecomputelatent}
Jonas Geiping, Sean McLeish, Neel Jain, John Kirchenbauer, Siddharth Singh, Brian~R. Bartoldson, Bhavya Kailkhura, Abhinav Bhatele, and Tom Goldstein.
\newblock Scaling up test-time compute with latent reasoning: A recurrent depth approach, 2025.
\newblock URL \url{https://arxiv.org/abs/2502.05171}.

\bibitem[Gou et~al.(2024)Gou, Shao, Gong, Shen, Yang, Duan, and Chen]{gou2024criticlargelanguagemodels}
Zhibin Gou, Zhihong Shao, Yeyun Gong, Yelong Shen, Yujiu Yang, Nan Duan, and Weizhu Chen.
\newblock Critic: Large language models can self-correct with tool-interactive critiquing, 2024.
\newblock URL \url{https://arxiv.org/abs/2305.11738}.

\bibitem[Grattafiori et~al.(2024)Grattafiori, Dubey, Jauhri, Pandey, Kadian, Al-Dahle, Letman, Mathur, Schelten, Vaughan, et~al.]{grattafiori2024llama}
Aaron Grattafiori, Abhimanyu Dubey, Abhinav Jauhri, Abhinav Pandey, Abhishek Kadian, Ahmad Al-Dahle, Aiesha Letman, Akhil Mathur, Alan Schelten, Alex Vaughan, et~al.
\newblock The llama 3 herd of models.
\newblock \emph{arXiv preprint arXiv:2407.21783}, 2024.

\bibitem[Gui et~al.(2024)Gui, Gârbacea, and Veitch]{gui2024bonbonalignmentlargelanguage}
Lin Gui, Cristina Gârbacea, and Victor Veitch.
\newblock Bonbon alignment for large language models and the sweetness of best-of-n sampling, 2024.
\newblock URL \url{https://arxiv.org/abs/2406.00832}.

\bibitem[Guo et~al.(2025)Guo, Yang, Zhang, Song, Zhang, Xu, Zhu, Ma, Wang, Bi, et~al.]{guo2025deepseek}
Daya Guo, Dejian Yang, Haowei Zhang, Junxiao Song, Ruoyu Zhang, Runxin Xu, Qihao Zhu, Shirong Ma, Peiyi Wang, Xiao Bi, et~al.
\newblock Deepseek-r1: Incentivizing reasoning capability in llms via reinforcement learning.
\newblock \emph{arXiv preprint arXiv:2501.12948}, 2025.

\bibitem[Hao et~al.(2024)Hao, Sukhbaatar, Su, Li, Hu, Weston, and Tian]{hao2024coconuttraininglargelanguagemodels}
Shibo Hao, Sainbayar Sukhbaatar, DiJia Su, Xian Li, Zhiting Hu, Jason Weston, and Yuandong Tian.
\newblock Training large language models to reason in a continuous latent space, 2024.
\newblock URL \url{https://arxiv.org/abs/2412.06769}.

\bibitem[He et~al.(2024)He, Chen, He, Yan, Wei, Luo, and Ling]{he2024retrievingrethinkingrevisingchainofverification}
Bolei He, Nuo Chen, Xinran He, Lingyong Yan, Zhenkai Wei, Jinchang Luo, and Zhen-Hua Ling.
\newblock Retrieving, rethinking and revising: The chain-of-verification can improve retrieval augmented generation, 2024.
\newblock URL \url{https://arxiv.org/abs/2410.05801}.

\bibitem[Hendrycks et~al.(2021{\natexlab{a}})Hendrycks, Burns, Basart, Zou, Mazeika, Song, and Steinhardt]{hendrycks2021measuringmassivemultitasklanguage}
Dan Hendrycks, Collin Burns, Steven Basart, Andy Zou, Mantas Mazeika, Dawn Song, and Jacob Steinhardt.
\newblock Measuring massive multitask language understanding, 2021{\natexlab{a}}.
\newblock URL \url{https://arxiv.org/abs/2009.03300}.

\bibitem[Hendrycks et~al.(2021{\natexlab{b}})Hendrycks, Burns, Kadavath, Arora, Basart, Tang, Song, and Steinhardt]{hendrycks2021measuringmathematicalproblemsolving}
Dan Hendrycks, Collin Burns, Saurav Kadavath, Akul Arora, Steven Basart, Eric Tang, Dawn Song, and Jacob Steinhardt.
\newblock Measuring mathematical problem solving with the math dataset, 2021{\natexlab{b}}.
\newblock URL \url{https://arxiv.org/abs/2103.03874}.

\bibitem[Hu et~al.(2025)Hu, Liu, Xu, and Shen]{hu2025reinforceefficientrlhfalgorithm}
Jian Hu, Jason~Klein Liu, Haotian Xu, and Wei Shen.
\newblock Reinforce++: An efficient rlhf algorithm with robustness to both prompt and reward models, 2025.
\newblock URL \url{https://arxiv.org/abs/2501.03262}.

\bibitem[H{\"u}botter et~al.(2025)H{\"u}botter, Bongni, Hakimi, and Krause]{hubotter2025efficientlysift}
Jonas H{\"u}botter, Sascha Bongni, Ido Hakimi, and Andreas Krause.
\newblock Efficiently learning at test-time: Active fine-tuning of {LLM}s.
\newblock In \emph{The Thirteenth International Conference on Learning Representations}, 2025.
\newblock URL \url{https://openreview.net/forum?id=NS1G1Uhny3}.

\bibitem[Irvine et~al.(2023)Irvine, Boubert, Raina, Liusie, Zhu, Mudupalli, Korshuk, Liu, Cremer, Assassi, Beauchamp, Lu, Rialan, and Beauchamp]{irvine2023rewardingchatbotsrealworldengagement}
Robert Irvine, Douglas Boubert, Vyas Raina, Adian Liusie, Ziyi Zhu, Vineet Mudupalli, Aliaksei Korshuk, Zongyi Liu, Fritz Cremer, Valentin Assassi, Christie-Carol Beauchamp, Xiaoding Lu, Thomas Rialan, and William Beauchamp.
\newblock Rewarding chatbots for real-world engagement with millions of users, 2023.
\newblock URL \url{https://arxiv.org/abs/2303.06135}.

\bibitem[Kang et~al.(2025)Kang, Jeong, and Cho]{kang2025t1toolintegratedselfverificationtesttime}
Minki Kang, Jongwon Jeong, and Jaewoong Cho.
\newblock T1: Tool-integrated self-verification for test-time compute scaling in small language models, 2025.
\newblock URL \url{https://arxiv.org/abs/2504.04718}.

\bibitem[Kaplan et~al.(2020)Kaplan, McCandlish, Henighan, Brown, Chess, Child, Gray, Radford, Wu, and Amodei]{kaplan2020scalinglawsneurallanguage}
Jared Kaplan, Sam McCandlish, Tom Henighan, Tom~B. Brown, Benjamin Chess, Rewon Child, Scott Gray, Alec Radford, Jeffrey Wu, and Dario Amodei.
\newblock Scaling laws for neural language models, 2020.
\newblock URL \url{https://arxiv.org/abs/2001.08361}.

\bibitem[Kumaran et~al.(2016)Kumaran, Hassabis, and McClelland]{kumaran2016learning}
Dharshan Kumaran, Demis Hassabis, and James~L McClelland.
\newblock What learning systems do intelligent agents need? complementary learning systems theory updated.
\newblock \emph{Trends in cognitive sciences}, 20\penalty0 (7):\penalty0 512--534, 2016.

\bibitem[Lee et~al.(2025)Lee, Fischer, Wu, Marwood, Baluja, Schuurmans, and Chen]{lee2025evolvingdeeperllmthinking}
Kuang-Huei Lee, Ian Fischer, Yueh-Hua Wu, Dave Marwood, Shumeet Baluja, Dale Schuurmans, and Xinyun Chen.
\newblock Evolving deeper llm thinking, 2025.
\newblock URL \url{https://arxiv.org/abs/2501.09891}.

\bibitem[Li et~al.(2025{\natexlab{a}})Li, Li, Wu, Zhu, Wang, Yu, Jiang, Zhu, Jia, Wu, and Zheng]{li2025seekdarkreasoningtesttime}
Hengli Li, Chenxi Li, Tong Wu, Xuekai Zhu, Yuxuan Wang, Zhaoxin Yu, Eric~Hanchen Jiang, Song-Chun Zhu, Zixia Jia, Ying~Nian Wu, and Zilong Zheng.
\newblock Seek in the dark: Reasoning via test-time instance-level policy gradient in latent space, 2025{\natexlab{a}}.
\newblock URL \url{https://arxiv.org/abs/2505.13308}.

\bibitem[Li et~al.(2024)Li, Beeching, Tunstall, Lipkin, Soletskyi, Huang, Rasul, Yu, Jiang, Shen, et~al.]{li2024numinamath}
Jia Li, Edward Beeching, Lewis Tunstall, Ben Lipkin, Roman Soletskyi, Shengyi Huang, Kashif Rasul, Longhui Yu, Albert~Q Jiang, Ziju Shen, et~al.
\newblock Numinamath: The largest public dataset in ai4maths with 860k pairs of competition math problems and solutions.
\newblock \emph{Hugging Face repository}, 13\penalty0 (9):\penalty0 9, 2024.

\bibitem[Li et~al.(2025{\natexlab{b}})Li, Dong, Jin, Zhang, Zhou, Zhu, Zhang, and Dou]{li2025searcho1agenticsearchenhancedlarge}
Xiaoxi Li, Guanting Dong, Jiajie Jin, Yuyao Zhang, Yujia Zhou, Yutao Zhu, Peitian Zhang, and Zhicheng Dou.
\newblock Search-o1: Agentic search-enhanced large reasoning models, 2025{\natexlab{b}}.
\newblock URL \url{https://arxiv.org/abs/2501.05366}.

\bibitem[Lifshitz et~al.(2025)Lifshitz, McIlraith, and Du]{lifshitz2025multi}
Shalev Lifshitz, Sheila~A McIlraith, and Yilun Du.
\newblock Multi-agent verification: Scaling test-time compute with multiple verifiers.
\newblock \emph{arXiv preprint arXiv:2502.20379}, 2025.

\bibitem[Liu et~al.(2024)Liu, Pfeiffer, Wu, Xie, and Szlam]{liu2024deliberation}
Luyang Liu, Jonas Pfeiffer, Jiaxing Wu, Jun Xie, and Arthur Szlam.
\newblock Deliberation in latent space via differentiable cache augmentation.
\newblock \emph{arXiv preprint arXiv:2412.17747}, 2024.

\bibitem[Liu et~al.(2021)Liu, Kothari, Van~Delft, Bellot-Gurlet, Mordan, and Alahi]{liu2021ttt++}
Yuejiang Liu, Parth Kothari, Bastien Van~Delft, Baptiste Bellot-Gurlet, Taylor Mordan, and Alexandre Alahi.
\newblock Ttt++: When does self-supervised test-time training fail or thrive?
\newblock \emph{Advances in Neural Information Processing Systems}, 34:\penalty0 21808--21820, 2021.

\bibitem[Luo et~al.(2025)Luo, Jain, Singh, Tan, Patel, Wu, Ariyak, Cai, Tarun~Venkat, Athiwaratkun, Roongta, Zhang, Li, Popa, Sen, and Stoica]{deepswe2025}
Michael Luo, Naman Jain, Jaskirat Singh, Sijun Tan, Ameen Patel, Qingyang Wu, Alpay Ariyak, Colin Cai, Shang~Zhu Tarun~Venkat, Ben Athiwaratkun, Manan Roongta, Ce~Zhang, Li~Erran Li, Raluca~Ada Popa, Koushik Sen, and Ion Stoica.
\newblock Deepswe: Training a state-of-the-art coding agent from scratch by scaling rl.
\newblock \url{https://pretty-radio-b75.notion.site/DeepSWE-Training-a-Fully-Open-sourced-State-of-the-Art-Coding-Agent-by-Scaling-RL-22281902c1468193aabbe9a8c59bbe33}, 2025.
\newblock Notion Blog.

\bibitem[Madaan et~al.(2023)Madaan, Tandon, Gupta, Hallinan, Gao, Wiegreffe, Alon, Dziri, Prabhumoye, Yang, Gupta, Majumder, Hermann, Welleck, Yazdanbakhsh, and Clark]{madaan2023selfrefineiterativerefinementselffeedback}
Aman Madaan, Niket Tandon, Prakhar Gupta, Skyler Hallinan, Luyu Gao, Sarah Wiegreffe, Uri Alon, Nouha Dziri, Shrimai Prabhumoye, Yiming Yang, Shashank Gupta, Bodhisattwa~Prasad Majumder, Katherine Hermann, Sean Welleck, Amir Yazdanbakhsh, and Peter Clark.
\newblock Self-refine: Iterative refinement with self-feedback, 2023.
\newblock URL \url{https://arxiv.org/abs/2303.17651}.

\bibitem[McClelland et~al.(1995)McClelland, McNaughton, and O'Reilly]{mcclelland1995there}
James~L McClelland, Bruce~L McNaughton, and Randall~C O'Reilly.
\newblock Why there are complementary learning systems in the hippocampus and neocortex: insights from the successes and failures of connectionist models of learning and memory.
\newblock \emph{Psychological review}, 102\penalty0 (3):\penalty0 419, 1995.

\bibitem[Orlicki(2025)]{orlicki2025wordslatentmemoryapproach}
José~I. Orlicki.
\newblock Beyond words: A latent memory approach to internal reasoning in llms, 2025.
\newblock URL \url{https://arxiv.org/abs/2502.21030}.

\bibitem[Peng et~al.(2024)Peng, Wu, Wang, and Fang]{peng2024softprompttuningaugmenting}
Zhiyuan Peng, Xuyang Wu, Qifan Wang, and Yi~Fang.
\newblock Soft prompt tuning for augmenting dense retrieval with large language models, 2024.
\newblock URL \url{https://arxiv.org/abs/2307.08303}.

\bibitem[Qiu et~al.(2025{\natexlab{a}})Qiu, Juan, Wang, Yang, Qi, Zhang, Guo, Lu, Yao, Wang, Liu, Jiang, Leqi, and Wang]{qiu2025agentdistilltrainingfreeagentdistillation}
Jiahao Qiu, Xinzhe Juan, Yimin Wang, Ling Yang, Xuan Qi, Tongcheng Zhang, Jiacheng Guo, Yifu Lu, Zixin Yao, Hongru Wang, Shilong Liu, Xun Jiang, Liu Leqi, and Mengdi Wang.
\newblock Agentdistill: Training-free agent distillation with generalizable mcp boxes, 2025{\natexlab{a}}.
\newblock URL \url{https://arxiv.org/abs/2506.14728}.

\bibitem[Qiu et~al.(2025{\natexlab{b}})Qiu, Qi, Zhang, Juan, Guo, Lu, Wang, Yao, Ren, Jiang, Zhou, Liu, Yang, Wu, Huang, Liu, Wang, and Wang]{qiu2025alitageneralistagentenabling}
Jiahao Qiu, Xuan Qi, Tongcheng Zhang, Xinzhe Juan, Jiacheng Guo, Yifu Lu, Yimin Wang, Zixin Yao, Qihan Ren, Xun Jiang, Xing Zhou, Dongrui Liu, Ling Yang, Yue Wu, Kaixuan Huang, Shilong Liu, Hongru Wang, and Mengdi Wang.
\newblock Alita: Generalist agent enabling scalable agentic reasoning with minimal predefinition and maximal self-evolution, 2025{\natexlab{b}}.
\newblock URL \url{https://arxiv.org/abs/2505.20286}.

\bibitem[Qwen et~al.(2025)Qwen, :, Yang, Yang, Zhang, Hui, Zheng, Yu, Li, Liu, Huang, Wei, Lin, Yang, Tu, Zhang, Yang, Yang, Zhou, Lin, Dang, Lu, Bao, Yang, Yu, Li, Xue, Zhang, Zhu, Men, Lin, Li, Tang, Xia, Ren, Ren, Fan, Su, Zhang, Wan, Liu, Cui, Zhang, and Qiu]{qwen2025qwen25technicalreport}
Qwen, :, An~Yang, Baosong Yang, Beichen Zhang, Binyuan Hui, Bo~Zheng, Bowen Yu, Chengyuan Li, Dayiheng Liu, Fei Huang, Haoran Wei, Huan Lin, Jian Yang, Jianhong Tu, Jianwei Zhang, Jianxin Yang, Jiaxi Yang, Jingren Zhou, Junyang Lin, Kai Dang, Keming Lu, Keqin Bao, Kexin Yang, Le~Yu, Mei Li, Mingfeng Xue, Pei Zhang, Qin Zhu, Rui Men, Runji Lin, Tianhao Li, Tianyi Tang, Tingyu Xia, Xingzhang Ren, Xuancheng Ren, Yang Fan, Yang Su, Yichang Zhang, Yu~Wan, Yuqiong Liu, Zeyu Cui, Zhenru Zhang, and Zihan Qiu.
\newblock Qwen2.5 technical report, 2025.
\newblock URL \url{https://arxiv.org/abs/2412.15115}.

\bibitem[Rein et~al.(2023)Rein, Hou, Stickland, Petty, Pang, Dirani, Michael, and Bowman]{rein2023gpqagraduatelevelgoogleproofqa}
David Rein, Betty~Li Hou, Asa~Cooper Stickland, Jackson Petty, Richard~Yuanzhe Pang, Julien Dirani, Julian Michael, and Samuel~R. Bowman.
\newblock Gpqa: A graduate-level google-proof qa benchmark, 2023.
\newblock URL \url{https://arxiv.org/abs/2311.12022}.

\bibitem[Shinn et~al.(2023)Shinn, Cassano, Berman, Gopinath, Narasimhan, and Yao]{shinn2023reflexionlanguageagentsverbal}
Noah Shinn, Federico Cassano, Edward Berman, Ashwin Gopinath, Karthik Narasimhan, and Shunyu Yao.
\newblock Reflexion: Language agents with verbal reinforcement learning, 2023.
\newblock URL \url{https://arxiv.org/abs/2303.11366}.

\bibitem[Snell et~al.(2024)Snell, Lee, Xu, and Kumar]{snell2024scalingllmtesttimecompute}
Charlie Snell, Jaehoon Lee, Kelvin Xu, and Aviral Kumar.
\newblock Scaling llm test-time compute optimally can be more effective than scaling model parameters, 2024.
\newblock URL \url{https://arxiv.org/abs/2408.03314}.

\bibitem[Song et~al.(2024)Song, Xiong, Zhao, Zhu, Wu, Wang, Li, Peng, and Li]{song2024agentbankgeneralizedllmagents}
Yifan Song, Weimin Xiong, Xiutian Zhao, Dawei Zhu, Wenhao Wu, Ke~Wang, Cheng Li, Wei Peng, and Sujian Li.
\newblock Agentbank: Towards generalized llm agents via fine-tuning on 50000+ interaction trajectories, 2024.
\newblock URL \url{https://arxiv.org/abs/2410.07706}.

\bibitem[Su et~al.(2025)Su, Wang, Ren, Lin, and Chen]{su2025pixelreasonerincentivizingpixelspace}
Alex Su, Haozhe Wang, Weiming Ren, Fangzhen Lin, and Wenhu Chen.
\newblock Pixel reasoner: Incentivizing pixel-space reasoning with curiosity-driven reinforcement learning, 2025.
\newblock URL \url{https://arxiv.org/abs/2505.15966}.

\bibitem[Sun et~al.(2020)Sun, Wang, Liu, Miller, Efros, and Hardt]{sun2020test}
Yu~Sun, Xiaolong Wang, Zhuang Liu, John Miller, Alexei Efros, and Moritz Hardt.
\newblock Test-time training with self-supervision for generalization under distribution shifts.
\newblock In \emph{International conference on machine learning}, pp.\  9229--9248. PMLR, 2020.

\bibitem[Sun et~al.(2025)Sun, Chen, Li, and Ding]{sun2025enhancinglatentcomputationtransformers}
Yuchang Sun, Yanxi Chen, Yaliang Li, and Bolin Ding.
\newblock Enhancing latent computation in transformers with latent tokens, 2025.
\newblock URL \url{https://arxiv.org/abs/2505.12629}.

\bibitem[Suzgun et~al.(2025)Suzgun, Yuksekgonul, Bianchi, Jurafsky, and Zou]{suzgun2025dynamiccheatsheettesttimelearning}
Mirac Suzgun, Mert Yuksekgonul, Federico Bianchi, Dan Jurafsky, and James Zou.
\newblock Dynamic cheatsheet: Test-time learning with adaptive memory, 2025.
\newblock URL \url{https://arxiv.org/abs/2504.07952}.

\bibitem[Tan et~al.(2025)Tan, Li, Ju, Luo, Luan, and Song]{tan2025thinksilentlythinkfast}
Wenhui Tan, Jiaze Li, Jianzhong Ju, Zhenbo Luo, Jian Luan, and Ruihua Song.
\newblock Think silently, think fast: Dynamic latent compression of llm reasoning chains, 2025.
\newblock URL \url{https://arxiv.org/abs/2505.16552}.

\bibitem[Tang et~al.(2025)Tang, Qin, Peng, Zhou, Shao, Du, Wei, Xia, Wu, Zhu, Zhang, Liu, Wang, Hong, Wu, Cheng, Wang, and Zhou]{tang2025agentkbleveragingcrossdomain}
Xiangru Tang, Tianrui Qin, Tianhao Peng, Ziyang Zhou, Daniel Shao, Tingting Du, Xinming Wei, Peng Xia, Fang Wu, He~Zhu, Ge~Zhang, Jiaheng Liu, Xingyao Wang, Sirui Hong, Chenglin Wu, Hao Cheng, Chi Wang, and Wangchunshu Zhou.
\newblock Agent kb: Leveraging cross-domain experience for agentic problem solving, 2025.
\newblock URL \url{https://arxiv.org/abs/2507.06229}.

\bibitem[Team et~al.(2025)Team, Kamath, Ferret, Pathak, Vieillard, Merhej, Perrin, Matejovicova, Ram{\'e}, Rivi{\`e}re, et~al.]{team2025gemma}
Gemma Team, Aishwarya Kamath, Johan Ferret, Shreya Pathak, Nino Vieillard, Ramona Merhej, Sarah Perrin, Tatiana Matejovicova, Alexandre Ram{\'e}, Morgane Rivi{\`e}re, et~al.
\newblock Gemma 3 technical report.
\newblock \emph{arXiv preprint arXiv:2503.19786}, 2025.

\bibitem[Villalobos et~al.(2022)Villalobos, Ho, Sevilla, Besiroglu, Heim, and Hobbhahn]{villalobos2022will}
Pablo Villalobos, Anson Ho, Jaime Sevilla, Tamay Besiroglu, Lennart Heim, and Marius Hobbhahn.
\newblock Will we run out of data? limits of llm scaling based on human-generated data.
\newblock \emph{arXiv preprint arXiv:2211.04325}, 2022.

\bibitem[Wang et~al.(2023{\natexlab{a}})Wang, Xie, Jiang, Mandlekar, Xiao, Zhu, Fan, and Anandkumar]{wang2023voyageropenendedembodiedagent}
Guanzhi Wang, Yuqi Xie, Yunfan Jiang, Ajay Mandlekar, Chaowei Xiao, Yuke Zhu, Linxi Fan, and Anima Anandkumar.
\newblock Voyager: An open-ended embodied agent with large language models, 2023{\natexlab{a}}.
\newblock URL \url{https://arxiv.org/abs/2305.16291}.

\bibitem[Wang et~al.(2024{\natexlab{a}})Wang, Wang, Athiwaratkun, Zhang, and Zou]{wang2024mixture}
Junlin Wang, Jue Wang, Ben Athiwaratkun, Ce~Zhang, and James Zou.
\newblock Mixture-of-agents enhances large language model capabilities.
\newblock \emph{arXiv preprint arXiv:2406.04692}, 2024{\natexlab{a}}.

\bibitem[Wang et~al.(2024{\natexlab{b}})Wang, Hu, Lu, Zhu, Zhang, Subramaniam, Loomba, Zhang, Sun, and Wang]{wang2024scibenchevaluatingcollegelevelscientific}
Xiaoxuan Wang, Ziniu Hu, Pan Lu, Yanqiao Zhu, Jieyu Zhang, Satyen Subramaniam, Arjun~R. Loomba, Shichang Zhang, Yizhou Sun, and Wei Wang.
\newblock Scibench: Evaluating college-level scientific problem-solving abilities of large language models, 2024{\natexlab{b}}.
\newblock URL \url{https://arxiv.org/abs/2307.10635}.

\bibitem[Wang et~al.(2023{\natexlab{b}})Wang, Wei, Schuurmans, Le, Chi, Narang, Chowdhery, and Zhou]{wang2023selfconsistencyimproveschainthought}
Xuezhi Wang, Jason Wei, Dale Schuurmans, Quoc Le, Ed~Chi, Sharan Narang, Aakanksha Chowdhery, and Denny Zhou.
\newblock Self-consistency improves chain of thought reasoning in language models, 2023{\natexlab{b}}.
\newblock URL \url{https://arxiv.org/abs/2203.11171}.

\bibitem[Wang et~al.(2024{\natexlab{c}})Wang, Gao, Chen, Jiang, Li, Yang, Yin, Li, Li, Yin, et~al.]{wang2024memoryllm}
Yu~Wang, Yifan Gao, Xiusi Chen, Haoming Jiang, Shiyang Li, Jingfeng Yang, Qingyu Yin, Zheng Li, Xian Li, Bing Yin, et~al.
\newblock Memoryllm: Towards self-updatable large language models.
\newblock \emph{arXiv preprint arXiv:2402.04624}, 2024{\natexlab{c}}.

\bibitem[Wang et~al.(2025{\natexlab{a}})Wang, Krotov, Hu, Gao, Zhou, McAuley, Gutfreund, Feris, and He]{wang2025mextendingmemoryllmscalable}
Yu~Wang, Dmitry Krotov, Yuanzhe Hu, Yifan Gao, Wangchunshu Zhou, Julian McAuley, Dan Gutfreund, Rogerio Feris, and Zexue He.
\newblock M+: Extending memoryllm with scalable long-term memory, 2025{\natexlab{a}}.
\newblock URL \url{https://arxiv.org/abs/2502.00592}.

\bibitem[Wang et~al.(2025{\natexlab{b}})Wang, Xu, Wang, Zhang, Yan, Zhang, Huang, and Ji]{wang2025mobileagenteselfevolvingmobileassistant}
Zhenhailong Wang, Haiyang Xu, Junyang Wang, Xi~Zhang, Ming Yan, Ji~Zhang, Fei Huang, and Heng Ji.
\newblock Mobile-agent-e: Self-evolving mobile assistant for complex tasks, 2025{\natexlab{b}}.
\newblock URL \url{https://arxiv.org/abs/2501.11733}.

\bibitem[Wei et~al.(2023)Wei, Wang, Schuurmans, Bosma, Ichter, Xia, Chi, Le, and Zhou]{wei2023cot}
Jason Wei, Xuezhi Wang, Dale Schuurmans, Maarten Bosma, Brian Ichter, Fei Xia, Ed~Chi, Quoc Le, and Denny Zhou.
\newblock Chain-of-thought prompting elicits reasoning in large language models, 2023.
\newblock URL \url{https://arxiv.org/abs/2201.11903}.

\bibitem[Wei et~al.(2025)Wei, Duchenne, Copet, Carbonneaux, Zhang, Fried, Synnaeve, Singh, and Wang]{wei2025swerladvancingllmreasoning}
Yuxiang Wei, Olivier Duchenne, Jade Copet, Quentin Carbonneaux, Lingming Zhang, Daniel Fried, Gabriel Synnaeve, Rishabh Singh, and Sida~I. Wang.
\newblock Swe-rl: Advancing llm reasoning via reinforcement learning on open software evolution, 2025.
\newblock URL \url{https://arxiv.org/abs/2502.18449}.

\bibitem[Williams(1992)]{williams1992simple}
Ronald~J Williams.
\newblock Simple statistical gradient-following algorithms for connectionist reinforcement learning.
\newblock \emph{Machine learning}, 8:\penalty0 229--256, 1992.

\bibitem[Wu et~al.(2025)Wu, Zhu, and Liu]{wu2025agenticreasoningreasoningllms}
Junde Wu, Jiayuan Zhu, and Yuyuan Liu.
\newblock Agentic reasoning: Reasoning llms with tools for the deep research, 2025.
\newblock URL \url{https://arxiv.org/abs/2502.04644}.

\bibitem[Xiang et~al.(2025)Xiang, Snell, Gandhi, Albalak, Singh, Blagden, Phung, Rafailov, Lile, Mahan, Castricato, Franken, Haber, and Finn]{xiang20252reasoningllmslearning}
Violet Xiang, Charlie Snell, Kanishk Gandhi, Alon Albalak, Anikait Singh, Chase Blagden, Duy Phung, Rafael Rafailov, Nathan Lile, Dakota Mahan, Louis Castricato, Jan-Philipp Franken, Nick Haber, and Chelsea Finn.
\newblock Towards system 2 reasoning in llms: Learning how to think with meta chain-of-thought, 2025.
\newblock URL \url{https://arxiv.org/abs/2501.04682}.

\bibitem[Xiao et~al.(2023)Xiao, Xu, Li, Lu, and Li]{xiao2023decomposedprompttuninglowrank}
Yao Xiao, Lu~Xu, Jiaxi Li, Wei Lu, and Xiaoli Li.
\newblock Decomposed prompt tuning via low-rank reparameterization, 2023.
\newblock URL \url{https://arxiv.org/abs/2310.10094}.

\bibitem[Xu et~al.(2025{\natexlab{a}})Xu, Yan, Ma, Zhao, Sun, Cheng, He, Liu, and Wu]{xu-etal-2025-genius}
Fangzhi Xu, Hang Yan, Chang Ma, Haiteng Zhao, Qiushi Sun, Kanzhi Cheng, Junxian He, Jun Liu, and Zhiyong Wu.
\newblock Genius: A generalizable and purely unsupervised self-training framework for advanced reasoning.
\newblock In Wanxiang Che, Joyce Nabende, Ekaterina Shutova, and Mohammad~Taher Pilehvar (eds.), \emph{Proceedings of the 63rd Annual Meeting of the Association for Computational Linguistics (Volume 1: Long Papers)}, pp.\  13153--13167, Vienna, Austria, July 2025{\natexlab{a}}. Association for Computational Linguistics.
\newblock ISBN 979-8-89176-251-0.
\newblock \doi{10.18653/v1/2025.acl-long.644}.
\newblock URL \url{https://aclanthology.org/2025.acl-long.644/}.

\bibitem[Xu et~al.(2025{\natexlab{b}})Xu, Guo, Zeng, and Miao]{xu2025softcotsoftchainofthoughtefficient}
Yige Xu, Xu~Guo, Zhiwei Zeng, and Chunyan Miao.
\newblock Softcot: Soft chain-of-thought for efficient reasoning with llms, 2025{\natexlab{b}}.
\newblock URL \url{https://arxiv.org/abs/2502.12134}.

\bibitem[Xu et~al.(2025{\natexlab{c}})Xu, Guo, Zeng, and Miao]{xu2025softcottesttimescalingsoft}
Yige Xu, Xu~Guo, Zhiwei Zeng, and Chunyan Miao.
\newblock Softcot++: Test-time scaling with soft chain-of-thought reasoning, 2025{\natexlab{c}}.
\newblock URL \url{https://arxiv.org/abs/2505.11484}.

\bibitem[Yang et~al.(2025)Yang, Li, Yang, Zhang, Hui, Zheng, Yu, Gao, Huang, Lv, et~al.]{yang2025qwen3}
An~Yang, Anfeng Li, Baosong Yang, Beichen Zhang, Binyuan Hui, Bo~Zheng, Bowen Yu, Chang Gao, Chengen Huang, Chenxu Lv, et~al.
\newblock Qwen3 technical report.
\newblock \emph{arXiv preprint arXiv:2505.09388}, 2025.

\bibitem[Yang et~al.(2024)Yang, Jimenez, Wettig, Lieret, Yao, Narasimhan, and Press]{yang2024sweagentagentcomputerinterfacesenable}
John Yang, Carlos~E. Jimenez, Alexander Wettig, Kilian Lieret, Shunyu Yao, Karthik Narasimhan, and Ofir Press.
\newblock Swe-agent: Agent-computer interfaces enable automated software engineering, 2024.
\newblock URL \url{https://arxiv.org/abs/2405.15793}.

\bibitem[Ye et~al.(2025)Ye, Lin, Ng, and Yan]{ye2025multiagentsamplingscalinginference}
Hai Ye, Mingbao Lin, Hwee~Tou Ng, and Shuicheng Yan.
\newblock Multi-agent sampling: Scaling inference compute for data synthesis with tree search-based agentic collaboration, 2025.
\newblock URL \url{https://arxiv.org/abs/2412.17061}.

\bibitem[Yu et~al.(2024)Yu, Xu, Weston, and Kulikov]{yu2024distilling21}
Ping Yu, Jing Xu, Jason Weston, and Ilia Kulikov.
\newblock Distilling system 2 into system 1, 2024.
\newblock URL \url{https://arxiv.org/abs/2407.06023}.

\bibitem[Yu et~al.(2025)Yu, Chen, Zhang, Tan, Zhu, Pang, Qian, Wang, Gururangan, Zhang, Kambadur, Mahajan, and Hou]{yu2025selfgeneratedcritiquesboostreward}
Yue Yu, Zhengxing Chen, Aston Zhang, Liang Tan, Chenguang Zhu, Richard~Yuanzhe Pang, Yundi Qian, Xuewei Wang, Suchin Gururangan, Chao Zhang, Melanie Kambadur, Dhruv Mahajan, and Rui Hou.
\newblock Self-generated critiques boost reward modeling for language models, 2025.
\newblock URL \url{https://arxiv.org/abs/2411.16646}.

\bibitem[Yuan et~al.(2025)Yuan, Pang, Cho, Li, Sukhbaatar, Xu, and Weston]{yuan2025selfrewardinglanguagemodels}
Weizhe Yuan, Richard~Yuanzhe Pang, Kyunghyun Cho, Xian Li, Sainbayar Sukhbaatar, Jing Xu, and Jason Weston.
\newblock Self-rewarding language models, 2025.
\newblock URL \url{https://arxiv.org/abs/2401.10020}.

\bibitem[Yue et~al.(2025)Yue, Zhang, Liu, Wan, Wang, Cheng, and Qi]{yue2025masrouterlearningroutellms}
Yanwei Yue, Guibin Zhang, Boyang Liu, Guancheng Wan, Kun Wang, Dawei Cheng, and Yiyan Qi.
\newblock Masrouter: Learning to route llms for multi-agent systems, 2025.
\newblock URL \url{https://arxiv.org/abs/2502.11133}.

\bibitem[Zeng et~al.(2023)Zeng, Liu, Lu, Wang, Liu, Dong, and Tang]{zeng2023agenttuningenablinggeneralizedagent}
Aohan Zeng, Mingdao Liu, Rui Lu, Bowen Wang, Xiao Liu, Yuxiao Dong, and Jie Tang.
\newblock Agenttuning: Enabling generalized agent abilities for llms, 2023.
\newblock URL \url{https://arxiv.org/abs/2310.12823}.

\bibitem[Zeng et~al.(2024)Zeng, Zhong, Zhao, Wei, Yang, He, Cheng, Hu, Liu, Yan, Fang, and Zhou]{zeng2024skyworkmathdatascalinglaws}
Liang Zeng, Liangjun Zhong, Liang Zhao, Tianwen Wei, Liu Yang, Jujie He, Cheng Cheng, Rui Hu, Yang Liu, Shuicheng Yan, Han Fang, and Yahui Zhou.
\newblock Skywork-math: Data scaling laws for mathematical reasoning in large language models -- the story goes on, 2024.
\newblock URL \url{https://arxiv.org/abs/2407.08348}.

\bibitem[Zhang et~al.(2025{\natexlab{a}})Zhang, Fu, Wan, Yu, Wang, and Yan]{zhang2025gmemorytracinghierarchicalmemory}
Guibin Zhang, Muxin Fu, Guancheng Wan, Miao Yu, Kun Wang, and Shuicheng Yan.
\newblock G-memory: Tracing hierarchical memory for multi-agent systems, 2025{\natexlab{a}}.
\newblock URL \url{https://arxiv.org/abs/2506.07398}.

\bibitem[Zhang et~al.(2025{\natexlab{b}})Zhang, Hu, Lu, Lange, and Clune]{zhang2025darwingodelmachineopenended}
Jenny Zhang, Shengran Hu, Cong Lu, Robert Lange, and Jeff Clune.
\newblock Darwin godel machine: Open-ended evolution of self-improving agents, 2025{\natexlab{b}}.
\newblock URL \url{https://arxiv.org/abs/2505.22954}.

\bibitem[Zhang et~al.(2025{\natexlab{c}})Zhang, Lyu, Sun, Wang, Zhang, Hua, Wu, Guo, Wang, Muennighoff, King, Liu, and Ma]{zhang2025surveytesttimescalinglarge}
Qiyuan Zhang, Fuyuan Lyu, Zexu Sun, Lei Wang, Weixu Zhang, Wenyue Hua, Haolun Wu, Zhihan Guo, Yufei Wang, Niklas Muennighoff, Irwin King, Xue Liu, and Chen Ma.
\newblock A survey on test-time scaling in large language models: What, how, where, and how well?, 2025{\natexlab{c}}.
\newblock URL \url{https://arxiv.org/abs/2503.24235}.

\bibitem[Zhang et~al.(2025{\natexlab{d}})Zhang, Li, Wang, Chen, Zhang, Ye, Feng, Wang, Wang, Wang, et~al.]{zhang2025avengers}
Yiqun Zhang, Hao Li, Chenxu Wang, Linyao Chen, Qiaosheng Zhang, Peng Ye, Shi Feng, Daling Wang, Zhen Wang, Xinrun Wang, et~al.
\newblock The avengers: A simple recipe for uniting smaller language models to challenge proprietary giants.
\newblock \emph{arXiv preprint arXiv:2505.19797}, 2025{\natexlab{d}}.

\bibitem[Zhao et~al.(2024)Zhao, Huang, Xu, Lin, Liu, and Huang]{zhao2024expelllmagentsexperiential}
Andrew Zhao, Daniel Huang, Quentin Xu, Matthieu Lin, Yong-Jin Liu, and Gao Huang.
\newblock Expel: Llm agents are experiential learners, 2024.
\newblock URL \url{https://arxiv.org/abs/2308.10144}.

\bibitem[Zhao et~al.(2025)Zhao, Wu, Yue, Wu, Xu, Yue, Lin, Wang, Wu, Zheng, and Huang]{zhao2025absolutezeroreinforcedselfplay}
Andrew Zhao, Yiran Wu, Yang Yue, Tong Wu, Quentin Xu, Yang Yue, Matthieu Lin, Shenzhi Wang, Qingyun Wu, Zilong Zheng, and Gao Huang.
\newblock Absolute zero: Reinforced self-play reasoning with zero data, 2025.
\newblock URL \url{https://arxiv.org/abs/2505.03335}.

\bibitem[Zheng et~al.(2025{\natexlab{a}})Zheng, Fatemi, Jin, Wang, Gandhi, Song, Gu, Srinivasa, Liu, Neubig, and Su]{zheng2025skillweaverwebagentsselfimprove}
Boyuan Zheng, Michael~Y. Fatemi, Xiaolong Jin, Zora~Zhiruo Wang, Apurva Gandhi, Yueqi Song, Yu~Gu, Jayanth Srinivasa, Gaowen Liu, Graham Neubig, and Yu~Su.
\newblock Skillweaver: Web agents can self-improve by discovering and honing skills, 2025{\natexlab{a}}.
\newblock URL \url{https://arxiv.org/abs/2504.07079}.

\bibitem[Zheng et~al.(2025{\natexlab{b}})Zheng, Yang, Hong, Zhao, Xu, Yang, Shen, and Yu]{zheng2025deepeyesincentivizingthinkingimages}
Ziwei Zheng, Michael Yang, Jack Hong, Chenxiao Zhao, Guohai Xu, Le~Yang, Chao Shen, and Xing Yu.
\newblock Deepeyes: Incentivizing "thinking with images" via reinforcement learning, 2025{\natexlab{b}}.
\newblock URL \url{https://arxiv.org/abs/2505.14362}.

\bibitem[Zhou et~al.(2025)Zhou, He, Zhou, Chen, Tang, Zhao, Tong, Li, Chen, Zhou, Sun, Hui, Wang, He, Liu, Zhou, and Wu]{zhou2025surveyllmtimesdata}
Xuanhe Zhou, Junxuan He, Wei Zhou, Haodong Chen, Zirui Tang, Haoyu Zhao, Xin Tong, Guoliang Li, Youmin Chen, Jun Zhou, Zhaojun Sun, Binyuan Hui, Shuo Wang, Conghui He, Zhiyuan Liu, Jingren Zhou, and Fan Wu.
\newblock A survey of llm $\times$ data, 2025.
\newblock URL \url{https://arxiv.org/abs/2505.18458}.

\bibitem[Zhu et~al.(2025)Zhu, Peng, Cheng, Qu, Huang, Zhu, Wang, Xue, Zhang, Shan, et~al.]{zhu2025surveylatentreasoning}
Rui-Jie Zhu, Tianhao Peng, Tianhao Cheng, Xingwei Qu, Jinfa Huang, Dawei Zhu, Hao Wang, Kaiwen Xue, Xuanliang Zhang, Yong Shan, et~al.
\newblock A survey on latent reasoning.
\newblock \emph{arXiv preprint arXiv:2507.06203}, 2025.

\bibitem[Zuo et~al.(2025)Zuo, Zhang, Sheng, Qu, Cui, Zhu, Li, Zhang, Long, Hua, Qi, Sun, Ma, Yuan, Ding, and Zhou]{zuo2025ttrltesttimereinforcementlearning}
Yuxin Zuo, Kaiyan Zhang, Li~Sheng, Shang Qu, Ganqu Cui, Xuekai Zhu, Haozhan Li, Yuchen Zhang, Xinwei Long, Ermo Hua, Biqing Qi, Youbang Sun, Zhiyuan Ma, Lifan Yuan, Ning Ding, and Bowen Zhou.
\newblock Ttrl: Test-time reinforcement learning, 2025.
\newblock URL \url{https://arxiv.org/abs/2504.16084}.

\end{thebibliography}
\bibliographystyle{iclr2025_conference}

\appendix

\section{Methodology Details}\label{app:method}

\subsection{Self-supervised Latent Refinement}\label{app:refine-latent}
The self-rewarding function $Q(\mathbf{y})$ in \Cref{eq:daytime-optimize-loss} is formally defined as a weighted aggregation of numerical scores assigned by the LLM to distinct evaluation criteria. Following standard practices from \citep{lifshitz2025multi,li2025seekdarkreasoningtesttime}, 
for each candidate output $\mathbf{y}$, the LLM produces normalized scores $s_{\text{ans}}(\mathbf{y}), s_{\text{comp}}(\mathbf{y}), s_{\text{calc}}(\mathbf{y}), s_{\text{form}}(\mathbf{y}), s_{\text{clar}}(\mathbf{y}) \in [0,1]$, corresponding respectively to (i) correctness of the final answer, (ii) accuracy of problem comprehension, (iii) validity of numerical calculations, (iv) conformity of the answer format to task requirements, and (v) provision of a clear and explicit answer.  
The overall reward is then computed as  
\begin{equation}
Q(\mathbf{y}) \;=\; \frac{1}{7}\Big( s_{\text{ans}}(\mathbf{y}) + s_{\text{comp}}(\mathbf{y}) + s_{\text{calc}}(\mathbf{y}) + 2\,s_{\text{form}}(\mathbf{y}) + 2\,s_{\text{clar}}(\mathbf{y}) \Big),
\end{equation}
where the weighting scheme $1:1:1:2:2$ reflects the relative importance of the criteria, placing greater emphasis on answer format fidelity and clarity of presentation.

\begin{tcolorbox}[notitle, sharp corners, breakable, colframe=Periwinkle, colback=white, 
       boxrule=3pt, boxsep=0.5pt, enhanced, 
       shadow={3pt}{-3pt}{0pt}{opacity=1,mygrey},
       title={Prompt for (i) correctness of the final answer},]
       {\scriptsize
\begin{lstlisting}
prompt_s_ans = f"""
INSTRUCTIONS:
Your task is to determine the correctness of the final answer within the PROPOSED SOLUTION.
Critically verify the final answer against the TASK DESCRIPTION and the reasoning steps provided in the PROPOSED SOLUTION.
Do NOT use external knowledge. Focus only on internal consistency and accuracy based on the given problem.

Your response must strictly follow the required format:
SCORE: [0.0-1.0]
(0.0 = completely incorrect, 1.0 = perfectly correct)

TASK DESCRIPTION:
[TASK_DESCRIPTION]

PROPOSED SOLUTION:
[PROPOSED_SOLUTION]
"""
\end{lstlisting}
}
\end{tcolorbox}

\begin{tcolorbox}[notitle, sharp corners, breakable, colframe=Periwinkle, colback=white, 
       boxrule=3pt, boxsep=0.5pt, enhanced, 
       shadow={3pt}{-3pt}{0pt}{opacity=1,mygrey},
       title={Prompt for (ii) accuracy of problem comprehension},]
       {\scriptsize
\begin{lstlisting}
prompt_s_comp = f"""
INSTRUCTIONS:
Your task is to evaluate the PROPOSED SOLUTION's understanding of the TASK DESCRIPTION.
Identify all explicit and implicit constraints, conditions, and specific requests in the TASK DESCRIPTION.
Assess how accurately and comprehensively the PROPOSED SOLUTION addressed these elements, demonstrating full comprehension.

Your response must strictly follow the required format:
SCORE: [0.0-1.0]
(0.0 = no comprehension, 1.0 = full and accurate comprehension)

TASK DESCRIPTION:
[TASK_DESCRIPTION]

PROPOSED SOLUTION:
[PROPOSED_SOLUTION]
"""
\end{lstlisting}
}
\end{tcolorbox}

\begin{tcolorbox}[notitle, sharp corners, breakable, colframe=Periwinkle, colback=white, 
       boxrule=3pt, boxsep=0.5pt, enhanced, 
       shadow={3pt}{-3pt}{0pt}{opacity=1,mygrey},
       title={Prompt for (iii) validity of numerical calculations},]
       {\scriptsize
\begin{lstlisting}
prompt_s_calc = f"""
INSTRUCTIONS:
Your task is to verify the validity of all numerical calculations and logical steps within the PROPOSED SOLUTION.
For each calculation or logical transition, independently recompute or re-evaluate it.
If any numerical or logical discrepancy is found, it indicates an error.

Your response must strictly follow the required format:
SCORE: [0.0-1.0]
(0.0 = many errors, 1.0 = all calculations and logical steps are valid)

TASK DESCRIPTION:
[TASK_DESCRIPTION]

PROPOSED SOLUTION:
[PROPOSED_SOLUTION]
"""
\end{lstlisting}
}
\end{tcolorbox}

\begin{tcolorbox}[notitle, sharp corners, breakable, colframe=Periwinkle, colback=white, 
       boxrule=3pt, boxsep=0.5pt, enhanced, 
       shadow={3pt}{-3pt}{0pt}{opacity=1,mygrey},
       title={Prompt for (iv) conformity of the answer format to task requirements},]
       {\scriptsize
\begin{lstlisting}
prompt_s_form = f"""
INSTRUCTIONS:
Your task is to assess if the PROPOSED SOLUTION conforms to the expected output format requirements.
Consider if specific units are used, if the answer is structured as implicitly or explicitly requested (e.g., numeric only, step-by-step, \\boxed{} formatting), and if all parts of the response are appropriately presented.

Your response must strictly follow the required format:
SCORE: [0.0-1.0]
(0.0 = completely incorrect format, 1.0 = perfectly formatted)

TASK DESCRIPTION:
[TASK_DESCRIPTION]

PROPOSED SOLUTION:
[PROPOSED_SOLUTION]
"""
\end{lstlisting}
}
\end{tcolorbox}

\begin{tcolorbox}[notitle, sharp corners, breakable, colframe=Periwinkle, colback=white, 
       boxrule=3pt, boxsep=0.5pt, enhanced, 
       shadow={3pt}{-3pt}{0pt}{opacity=1,mygrey},
       title={Prompt for (v) provision of a clear and explicit answer},]
       {\scriptsize
\begin{lstlisting}
prompt_s_clar = f"""
INSTRUCTIONS:
Your task is to evaluate the clarity and explicitness of the PROPOSED SOLUTION.
Assess if the reasoning is easy to follow, unambiguous, and if all necessary steps and explanations are provided without missing information.
Consider the overall readability and conciseness.

Your response must strictly follow the required format:
SCORE: [0.0-1.0]
(0.0 = very unclear/implicit, 1.0 = exceptionally clear and explicit)

TASK DESCRIPTION:
[TASK_DESCRIPTION]

PROPOSED SOLUTION:
[PROPOSED_SOLUTION]
"""
\end{lstlisting}
}
\end{tcolorbox}

\subsection{Nighttime Consolidation}\label{app:night}

When preparing training data for the latent weaver, not all encountered latent representations within a given cycle are equally valuable or should be retained. Recall that the rationale behind self-supervised refinement is as follows: although the latent state $\mathbf{z}_{0,i}$, obtained through weighted momentum transfer, may lie within a promising region of the space, it is not guaranteed to align perfectly with the current context. To address this, we further scale and update it through a self-rewarding mechanism, ensuring that it becomes optimally adapted to support the query at hand.
Therefore, the latent weaver, which provides the initial seeds for reasoning, should itself be trained with relatively high-quality data. Concretely, for a triplet $(\mathbf{e}_{\mathbf{c}_i}, \mathbf{z}_{0,i}, \mathbf{z}^*_i)$, we include it into the memory $\mathcal{M}$ if and only if the LLM exhibits sufficient confidence in the associated latent representation. Formally, such confidence is defined as the expected quality score of the final output after iterative refinement. Let $\mathbf{y}^{(M)}_{k}$ denote the response generated at the last refinement step under rollout $k$ $(k=1,\dots,M)$, then the confidence measure is given by
\[
\mathbb{E}[Q(\mathbf{y}^{(M)}_k)] = \frac{1}{M} \sum_{k=1}^M Q(\mathbf{y}^{(M)}_k),
\]
and the triplet is retained only if
\[
\mathbb{E}[Q(\mathbf{y}^{(M)}_k)] \geq \tau,
\]
where $Q(\mathbf{y}^{(M)}_k) \in [0,1]$ denotes the numerical score assigned to the generated response according to task-specific evaluation criteria, and $\tau$ is a tunable threshold that governs the admission of latent experiences. We set $\tau = 0.5$ across all experiments.

\subsection{Evaluation}\label{app:eval}

The evaluation prompts used by \ourmethod for datasets requiring numerical answers (including GSM8K, MATH, AIME 2024/2025, and SciBench) and for multiple-choice datasets (including MMLU, SciBench, JAMA, and GPQA) are summarized in \Cref{tab:eval-prompt}.

\begin{table*}[htbp]
\centering
\scriptsize
\ttfamily
\caption{Evaluation prompts for \ourmethod and other baselines.}
\vspace{-0.4em}
\label{tab:eval-prompt}
\begin{tabularx}{\textwidth}{X}
\toprule
\textbf{Numerical-answer evaluation prompt}: \{Question Description\}. Please reason step by step, and enclose your final answer within \(\backslash\backslash\)boxed\{\}.\\
\midrule
\textbf{Multiple-choice evaluation prompt}: \{Question Description\}. Please select the correct option (A, B, C, or D) to answer the question. Your response should be formatted as follows: The correct answer is \{your answer option letter here\}.\\
\bottomrule
\end{tabularx}
\end{table*}

\subsection{Dataset Details}\label{app:data}

This section provides the fine-grained statistics of each dataset:  
\begin{itemize}
    \item \textbf{MMLU}~\citep{hendrycks2021measuringmassivemultitasklanguage}: following prior practice~\citep{yue2025masrouterlearningroutellms}, we sample 1000 instances.  
    \item \textbf{MATH}~\citep{hendrycks2021measuringmathematicalproblemsolving}: we adopt the standard MATH-500 subset.  
    \item \textbf{GSM8K}~\citep{cobbe2021trainingverifier}: we opt for the full test set (1319 problems).  
    \item \textbf{GPQA}~\citep{rein2023gpqagraduatelevelgoogleproofqa}: we employ the GPQA-Diamond subset containing 198 graduate-level questions of elevated difficulty.  
    \item \textbf{SciBench}~\citep{wang2024scibenchevaluatingcollegelevelscientific}: we include all 692 tasks.  
    \item \textbf{JAMA Clinical Challenge}~\citep{chen2025benchmarkingjama}: comprising questions derived from demanding clinical cases, we adopt all 1511 test items.  
    \item \textbf{AIME 2024 and 2025}~\citep{li2024numinamath}: each consists of 30 problems.  
\end{itemize}


\subsection{More Results}\label{app:result}

\begin{table*}[ht]
\centering
\small
\caption{\textbf{Algorithmic Statistics} of \ourmethod. We report the ratio of answer length (when equipped with \ourmethod) to vanilla CoT length, under three settings (full \ourmethod, \textit{w/o Daytime} and \textit{w/o Nighttime}).}
\label{tab:algorithmic_statistics_new_llms}
\begin{tabular}{clccccc}
\toprule
\multicolumn{2}{c}{\textbf{Model}} &
  \begin{tabular}[c]{@{}c@{}}Qwen2.5\\ 7b\end{tabular} &
  \begin{tabular}[c]{@{}c@{}}Qwen3\\ 8b\end{tabular} &
  \begin{tabular}[c]{@{}c@{}}Qwen3\\ 4b\end{tabular} &
  \begin{tabular}[c]{@{}c@{}}Llama3.2\\ 3b\end{tabular} &
  \textbf{Avg} \\ 
\midrule
\multirow{3}{*}{\textbf{GSM8K}}    & \ourmethod  & 0.89 & 0.92 & 0.91 & 0.96 & 0.92 \\
                                   & \textit{w/o Daytime}   & 0.92 & 0.92 & 0.90 & 0.97 & 0.93 \\
                                   & \textit{w/o Nighttime} & 0.98 & 0.95 & 0.94 & 0.94 & 0.95 \\ \midrule
\multirow{3}{*}{\textbf{MATH-500}} & \ourmethod  & 0.91 & 0.93 & 0.95 & 0.97 & 0.94 \\
                                   & \textit{w/o Daytime}   & 0.93 & 0.93 & 1.01 & 0.99 & 0.97 \\
                                   & \textit{w/o Nighttime}  & 1.01 & 0.97 & 0.97 & 0.98 & 0.98 \\ \midrule
\multirow{3}{*}{\textbf{SciBench}} & \ourmethod & 0.93 & 0.95 & 0.94 & 0.98 & 0.95 \\
                                   & \textit{w/o Daytime}   & 0.94 & 0.98 & 0.95 & 1.04 & 0.98 \\
                                   & \textit{w/o Nighttime} & 0.99 & 0.95 & 0.92 & 1.02 & 0.97 \\ \bottomrule
\end{tabular}
\end{table*}

\subsection{Sensitivity Analysis}\label{app:sensi}

From \Cref{tab:iter-results}, we observe a clear improvement in performance with iterative refinement under moderate evolution intervals. For example, when $T=200$, accuracy steadily increases from $68.3\%$ at Iter~1 to $73.6\%$ at Iter~5, indicating that frequent daytime and nighttime interactions allow the latent representations to be progressively aligned with the query context. A similar trend is seen for $T=300$ and $T=500$, although the performance gains diminish as the interval grows. In contrast, when $T=1000$, the evolution becomes excessively coarse-grained, essentially reducing the process to a single daytime and nighttime interaction. This hinders the model’s ability to perform gradual refinement, thereby limiting the benefits of the proposed dual-phase evolution. These results suggest that overly sparse consolidation undermines the advantages of iterative scaling, while moderate intervals strike a balance between stability and adaptability.

\begin{table}[htbp]
\centering
\caption{Results across different iterations with different $T$ on the MMLU dataset ($T$ denotes the evolution interval controlling the frequency of daytime and nighttime interactions). 
Smaller $T$ values allow more iterative refinements within the same training budget, leading to smoother convergence, 
while larger $T$ values reduce the number of available iterations, coarsening the evolution process.}
\label{tab:iter-results}
\begin{tabular}{c|ccccc}
\toprule
 & \textbf{Iter 1} & \textbf{Iter 2} & \textbf{Iter 3} & \textbf{Iter 4} & \textbf{Iter 5} \\
\midrule
\textbf{200} & 68.3 & 70.2 & 72.8 & 73.3 & 73.6 \\
\textbf{300} & 70.2 & 71.9 & 73.3  & 73.3 & - \\
\textbf{500} & 72.1 & 73.9 & - & - & - \\
\textbf{1000} & 72.8 & - & - & - & - \\
\bottomrule
\end{tabular}
\end{table}

\end{document}